\documentclass[journal]{IEEEtran}

\ifCLASSINFOpdf
\else
\fi
\usepackage{graphicx}
\usepackage{epsfig}
\usepackage{epstopdf}
\usepackage{amsmath}
\usepackage{amssymb}
\usepackage{array}
\usepackage{color}
\usepackage{url}
\usepackage{multirow}
\usepackage{mathtools}
\usepackage{algorithm}
\usepackage{algorithmic}
\usepackage{diagbox}
\usepackage{makecell}
\usepackage{float}
\usepackage{subfig}

\newcommand{\ie}{{i.e., }}

\begin{document}
	%
	\title{Single-shot Phase Retrieval from \\ a Fractional Fourier Transform Perspective}
	%
	%
	%
	
	\author{Yixiao Yang,
		Ran Tao,~\IEEEmembership{Senior Member,~IEEE,}
		Kaixuan Wei,
        and Jun Shi
            \thanks{This work is supported in part by the National Natural Science Foundation of China under Grant 62027801 and Grant 61421001. (Corresponding author: Ran Tao.)}
		\thanks{Yixiao Yang and Ran Tao are with the Department of Information and Electronics, Beijing Institute of Technology, Beijing 100081, China (e-mail: yixiaoyang@bit.edu.cn and rantao@bit.edu.cn).}
		\thanks{Kaixuan Wei is with the Department of Computer Science, Princeton University, USA (e-mail:kxwei@princeton.edu).}
            \thanks{Jun Shi is with the Communication Research Center, Harbin Institute of Technology, Harbin 150001, China (e-mail:junshi@hit.edu.cn).}
	}
	
	\markboth{submitted to IEEE TRANSACTIONS ON SIGNAL PROCESSING}%
	{Shell \MakeLowercase{\textit{et al.}}: Bare Demo of IEEEtran.cls for IEEE Journals}

	\maketitle
	
	\begin{abstract}
        The realm of classical phase retrieval concerns itself with the arduous task of recovering a signal from its Fourier magnitude measurements, which are fraught with inherent ambiguities. A single-exposure intensity measurement is commonly deemed insufficient for the reconstruction of the primal signal, given that the absent phase component is imperative for the inverse transformation. In this work, we present a novel single-shot phase retrieval paradigm from a fractional Fourier transform (FrFT) perspective, which involves integrating the FrFT-based physical measurement model within a self-supervised reconstruction scheme. Specifically, the proposed FrFT-based measurement model addresses the aliasing artifacts problem in the numerical calculation of Fresnel diffraction, featuring adaptability to both short-distance and long-distance propagation scenarios. Moreover, the intensity measurement in the FrFT domain proves highly effective in alleviating the ambiguities of phase retrieval and relaxing the previous conditions on oversampled or multiple measurements in the Fourier domain. Furthermore, the proposed self-supervised reconstruction approach harnesses the fast discrete algorithm of FrFT alongside untrained neural network priors, thereby attaining preeminent results. Through numerical simulations, we demonstrate that both amplitude and phase objects can be effectively retrieved from a single-shot intensity measurement using the proposed approach and provide a promising technique for support-free coherent diffraction imaging.
	\end{abstract}
	
	
	%
	\IEEEpeerreviewmaketitle
	
	\section{Introduction}
	\label{sec:introduction}
	\IEEEPARstart{P}{hase} Retrieval (PR) is a long-established challenge for estimating a signal from the phase-less linear measurements, encountered in various science and engineering fields including x-ray crystallography \cite{miao1998phase}, computational microscopy \cite{zheng2013wide}, computer-generated holography \cite{chakravarthula2020learned}, and many more \cite{katz2014non}. In the realm of optical systems, the direct measurement of the phase by electronic detectors is often difficult, thus computational phase retrieval comes into play \cite{shechtman2015phase}. Furthermore, the observation in the far-field diffraction or the focal plane of a lens can be formulated by the Fourier transform, which enables Fourier PR.
	
    Despite its popularity, the Fourier PR problem faces a significant obstacle as it is not uniquely solvable solely from the Fourier magnitude \cite{bendory2017fourier}. This is due to the inherent structure of the Fourier transform, leading to trivial ambiguities such as target translation and inversion, as well as non-trivial ambiguities that preserve the Fourier magnitude. As a result, designing efficient and convergent reconstruction algorithms becomes an extremely challenging task. To overcome this issue, researchers have explored various strategies, such as obtaining additional measurements, incorporating prior knowledge of the object, or a combination of both.

    \begin{figure}[!t]
        \centering
        \includegraphics[width=0.9\linewidth,clip,keepaspectratio]{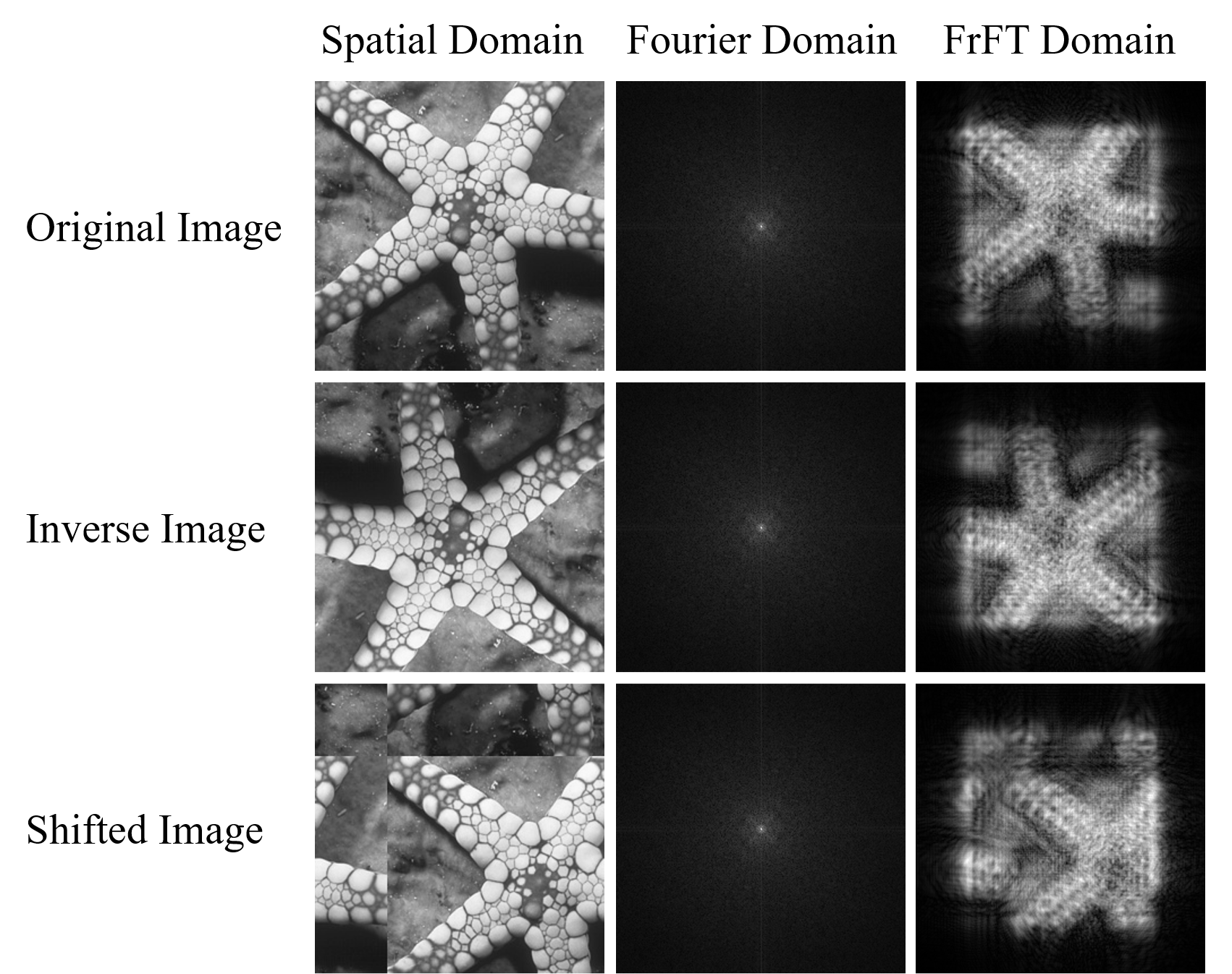}
        \caption{Operation of inversion or shift on the object corresponds to the same magnitude in the Fourier domain, but a different one in the FrFT domain. Here we adopt the $0.5$th-order fractional Fourier transform (FrFT).}
        \label{fig:trivial_ambiguities}
    \end{figure}
	

    \subsection{Related Work}
    Numerous physical measurement techniques have been developed over the years to record redundant information and serve as a remedy to the ill-posed nature of the PR problem. 
    A notable example is coherent diffraction imaging (CDI), which mitigates the ambiguities of PR by invoking pre-determined optical masks that offer valuable constraints \cite{zhang2016phase}. Pioneering research focused on utilizing the non-zero support as an optical mask, thereby enabling oversampled Fourier transform measurements and incorporating prior information on the signal, such as positivity and real-valuedness \cite{fienup1978reconstruction}.   
    Besides oversampling, the use of random masks has gained attention as a fashion to introduce multiple measurements within the optimization process \cite{candes2015cdp}. This technique is founded on the principle that various masks can modulate the signal of interest and introduce information redundancy to reduce ambiguities \cite{candes2015phase}. And it can be implemented through diverse means, including the utilization of masks \cite{seaberg2015coherent} and oblique illuminations \cite{faridian2010nanoscale}. 
    Another method to enrich observation diversity is ptychography or scanning CDI, which employs lateral sample (or probe) shifting to illuminate new regions of the sample while ensuring adequate overlap \cite{maiden2010optical}. The main idea is to seek uniqueness from the highly overlapped Fourier measurements \cite{jaganathan2016stft}.
    Despite the above successes, the requirement of redundant (oversampled/multiple/overlapped) measurements significantly increases system complexity and data acquisition time, therefore unsuitable for dynamic imaging scenarios \cite{takazawa2021demonstration}. In addition, in 3D imaging applications, it is extremely difficult to collect multiple observations and reliable single-shot PR methods are urgently needed \cite{chan2021rapid}.
    
    With the recent successes of deep learning techniques in computational imaging, there has been an increasing interest in employing data-driven methods for snapshot phase retrieval, whose goal is to retrieve the object from a single diffraction pattern. 
    This approach circumvents multiple measurements by leveraging labeled data to train a neural network to learn the inverse mapping of the single-shot measurement function.
    Notable examples of this class include works by Wang et al. \cite{wang2022snapshot}, as well as SiSPRNet \cite{ye2022sisprnet}. The former utilizes a trained convolutional neural network (CNN) to recover synthetic aperture images from a snapshot acquired on a 16-camera array, and the latter retrieves phase directly from a single Fourier-intensity measurement using CNN. 
    While these data-driven approaches yield promising results, they are limited in reduced interpretability and generalization as a result of solely relying on black-box neural networks and disregarding the underlying physical model. In addition, these supervised methods still face data challenges, requiring pairs of measurements and their corresponding ground truth images as training datasets. In many scientific imaging applications, collecting such paired data is extremely expensive if not impossible. 
    
    The above optical settings typically assume that measurements are taken at the Fourier plane or Fraunhofer regime. In fact, the intensity pattern can be collected at an arbitrary plane between the object field and the far field, implying the new measurement model that differs from Fourier transform. 
    The regime of Fresnel (near-field) diffraction, described by the Fresnel Integral, is the most notable case in this context. 
    This has led to growing interest in the concept of Fresnel Phase Retrieval \cite{li2008phase}, including Fresnel CDI \cite{williams2006fresnel} and Fresnel Ptychography \cite{iwen2023toward}.
    Although experimentally validated, it is susceptible to aliasing artifacts in the numerical calculation of near-field diffraction, arising from the sampling of non-band-limited chirp function \cite{zhang2020frequency}. Meanwhile, the Fractional Fourier Transform (FrFT), which employs the chirp function as its kernel, has garnered significant attention in the signal-processing community as a generalization of the Fourier transform \cite{frftoptics}. Developments in the theoretical framework of FrFT have been rapid \cite{jinming2018research}, covering sampling \cite{tao2007sampling}, filtering \cite{deng2006convolution}, and discrete algorithms \cite{su2019analysis}. In \cite{ozaktas1995fractional}, the definitional concept of fractional Fourier optics was introduced, based on the equivalent relationship between FrFT and Fresnel integral. This work revealed that the propagation of light between two spherical surfaces could be interpreted as a process of continual FrFT. Thus, this gives rise to a few works on the problem of phase retrieval from multiple FrFT magnitude measurements \cite{su2022phase}. Despite the theoretical merits, the existing research in this domain has predominantly disregarded the integration of quantitative analysis with numerical diffraction calculations. In addition, its potential in single-shot PR is largely overlooked by the community. 

    \subsection{Motivation and Contributions}
    To fill this gap, the work presented in this paper revisits the power of FrFT and proposes a novel single-shot PR paradigm from a joint physics and mathematics perspective with several insights developed. 
    Firstly, the FrFT, as a well-defined signal processing tool, can offer a new option for numerical diffraction calculations in the near field. Compared with the Fresnel integral, the FrFT benefits from the fractional Fourier sampling theory and fast discrete algorithms, avoiding aliasing artifacts in numerical calculations. Thus, the FrFT-based measurement model enables accurate and efficient computation of diffraction fields, bridging the gap between theory and practical applications. This finding supports that the single FrFT measurement can be accurately obtained through physical near-field diffraction, facilitating a snapshot operation without the need for extra optical components, such as masks.
    
    Secondly, the measurement in the FrFT domain has good and specific theoretical properties that could benefit PR. In contrast to the seriousness of the phase loss in the Fourier domain, the magnitude of the FrFT contains both amplitude and phase information of the original signal.
    Specifically, the FrFT provides a space-frequency representation of the signal, containing both spatial and frequency details. Therefore, changes in spatial signal amplitude are also reflected in the FrFT amplitude, which can be leveraged to overcome some spatial ambiguities of PR, as depicted in Fig.~\ref{fig:trivial_ambiguities}. In addition, the modulus of fractional Fourier space-frequency representation describes the frequency change of the signal in space and contains the signal's phase information. These theoretical foundations support the feasibility of recovering the original signal from a single FrFT measurement.
    
    Based on the above insights, the contributions of this paper are summarized in the following:
    \begin{itemize}
        \item To the best of our knowledge, we are the first to address the problem of reconstructing a two-dimensional image from the single intensity measurement in the FrFT domain, coined Single-shot Fractional Fourier Phase Retrieval (SFrFPR). 
        \item To this end, we formulate a detailed FrFT-based measurement model for near-field diffraction calculation, featuring adaptability to both short-distance and long-distance propagation scenarios. Furthermore, we propose a self-supervised reconstruction method, which harnesses the fast algorithm of FrFT alongside untrained neural network priors, thereby achieving superior results for recovering both amplitude and phase objects.
        \item Moreover, we provide theoretical analyses based on fractional Fourier time-frequency representation to clarify the rationality of the proposed SFrFPR. Through simulation, we demonstrate the single FrFT-based measurement effectively improves the uniqueness of the solution, relaxing the conditions on oversampled or multiple measurements in the Fourier domain. Last but not least, we present a promising imaging capability empowered by the proposed method, \ie support-free CDI.
    \end{itemize}
    
    \subsection{Outline}
    The remainder of this paper is organized as follows. Section \ref{sec:method} introduces the proposed SFrFPR, including the FrFT-based measurement model, self-supervised reconstructing approaches, and theoretical analysis. Numerical simulation results are presented in Section \ref{sec:experiments}. The conclusion is drawn in Section \ref{sec:conclusion}. 
    A limited version of this work with preliminary results was presented as a conference paper in 2023 IEEE ICASSP \cite{10095976}. In this manuscript, we further address key issues such as fast discrete algorithms, scalable sampling, and self-supervised reconstruction methods, rendering the proposed SFrFPR suitable for practical applications.
	
    \section{Method}
    \label{sec:method}

    In this section, we first recall the near-field diffraction theory and introduce the proposed FrFT-based measurement model. Then we formulate the SFrFPR problem and propose a self-supervised reconstructing approach based on an untrained neural network (UNN) scheme. Finally, theoretical analyses based on fractional Fourier time-frequency representation are provided.

    \subsection{Near-field Measurement Model based on FrFT}
    The proliferation of near-field measurements and imaging in recent years can be attributed to advancements in numerical propagation models. Instead of far-field measurement formulated by the Fourier transform, it is very challenging to establish a unified framework that can effectively, precisely, and flexibly compute the near-field diffraction. In this part, we present a novel near-field measurement model based on FrFT to solve this dilemma.

    \subsubsection{Near-field Diffraction}
    In the realm of optics, the Fresnel diffraction equation emerges as an approximation of the Kirchhoff–Fresnel diffraction formula, tailored for characterizing optical propagation in the near field. The diffraction pattern denoted as $U_d$ can be calculated when light traverses through an object aperture represented as $U_0$ by 
    \begin{equation}
        U_d(x,y) = \frac{e^{i\frac{2\pi}{\lambda} d}}{i\lambda d} \iint U_0(x',y')e^{i\frac{\pi}{\lambda d}[(x-x')^2+(y-y')^2]}dx'dy',
        \label{eq:fresnel}
    \end{equation}
    where $d$ denotes the propagation distance, $\lambda$ is the wavelength of light, and $i$ is the imaginary unit.

    Normally, this Fresnel integral \eqref{eq:fresnel} can be expressed by the single Fourier transform, called SFT-Fresnel, written as
    \begin{equation}
        U_d(x,y) = \frac{e^{i\frac{2\pi}{\lambda} d}}{i\lambda d} e^{i\frac{\pi (x^2+y^2)}{\lambda d}} \mathcal{F}[U_0(x',y')e^{i\frac{\pi (x'^2+y'^2)}{\lambda d}}](\frac{x}{\lambda d},\frac{y}{\lambda d}),
        \label{eq:sft}
    \end{equation}
    where $\mathcal{F}$ represents the two-dimensional Fourier transform.

    Note that there is a quadratic phase factor in the Fourier transform and that the phase oscillates rapidly over short propagation distances, which poses a great challenge for accurate sampling and computing. Even worse, when employing Fast Fourier Transform (FFT) for discrete calculations, it suffers from serious aliasing artifacts \cite{zhang2020frequency}.

    As a dual version, this Fresnel integral \eqref{eq:fresnel} can be also seen as a convolution operation \cite{li2007diffraction} and computed with the help of Fourier transform, stated as
    \begin{equation}
        U_d(x,y) = \mathcal{F}^{- 1}[\mathcal{F}[U_0(x',y')] \times H(f_x,f_y)],
        \label{eq:ftf}
    \end{equation}
    where $H(f_x,f_y) = \mathcal{F} [\frac{e^{i\frac{2\pi}{\lambda} d}}{i\lambda d} e^{i\frac{\pi (x^2+y^2)}{\lambda d}} ]$ denotes the transfer function of Fresnel diffraction and $(f_x,f_y)$ are the Fourier coordinates conjugate to the real space coordinates $(x,y)$. 
    
    Moreover, this transfer function has an analytical expression as $H(f_x,f_y) = e^{i\pi d(\frac{2}{\lambda} - \lambda (f_x^2 + f_y^2))}$.
    However, it is increasingly challenging to appropriately sample the transfer function, characterized by a swift oscillation of the phase component over long propagation distances. Numerical errors still exist when using FFT calculations duet to the fixed sampling pitch \cite{li2007diffraction}. While this problem can be solved by utilizing non-uniform FFT \cite{zhang2020band}, the computational complexity also increases, sacrificing the calculation efficiency.

    In summation, it is exceedingly arduous to formulate an efficient model that can compute the whole near-field diffraction without sampling problems. In the following, we will present an innovative FrFT-based model to address the sampling problem in the Fresnel integral without aliasing errors during the treatment of the chirp function, featuring adaptability to both short-distance and long-distance propagation scenarios.

    \subsubsection{FrFT-based Measurement model}
    The fractional Fourier transform (FrFT) is the generalized form of the Fourier transform, and its kernel function is a quadratic phase term.
    The $p$th-order FrFT of a continuous signal $f(x)$ is defined as \cite{almeida1994fractional}
    \begin{align}\label{FRFT-definition}
        X_\alpha(u)=\mathcal{F}^p [f](u)\triangleq \int f(x)K_\alpha(u,x)dx,
    \end{align}
    where $\mathcal{F}^p$ denotes the FrFT operator and $ K_\alpha(u,x) $ is the transform kernel with $\alpha = \frac{\pi}{2}p$ given as follows
    \begin{align}\label{FRFT kernel}
        K_\alpha(u,x)\triangleq
        \left\{
        \hspace{0pt}
    \begin{alignedat}{3}
        &A_\alpha e^{i\pi \left(\mathrm{cot}\alpha x^2- 2 \mathrm{csc}\alpha ux+\mathrm{cot} \alpha u^2\right)}, \;\alpha\neq n\pi \\
        &\delta(x-u), \;\alpha=2n\pi\\
        &\delta(x+u), \;\alpha =(2n\pm1)\pi
    \end{alignedat},
        \right.
    \end{align}
    with $A_\alpha$ defined as $A_\alpha\triangleq \sqrt{1-i\mathrm{cot}\alpha}$ and $\delta(t)$ being the Dirac delta function.
    Especially, the FrFT reduces to the (inverse) Fourier transform when $p = \pm 1$.
	
    Correspondingly, the inverse FrFT of $X_\alpha(u)$ in (\ref{FRFT-definition}) is
    \begin{align}\label{inverse-FRFT-definition}
        f(x)=\mathcal{F}^{-p}[X_\alpha](x)\triangleq \int X_\alpha(u)K_{-\alpha}(u,x)du,
    \end{align}
    where $\mathcal{F}^{-p}$ and $K_{-\alpha}(u,x)$ denote the inverse of $\mathcal{F}^p$ and the kernel $ K_\alpha(u,x) $ obtained from (\ref{FRFT-definition}) and (\ref{FRFT kernel}), respectively.

    \begin{figure}[!t]
        \centering
        \includegraphics[width=0.9\linewidth,clip,keepaspectratio]{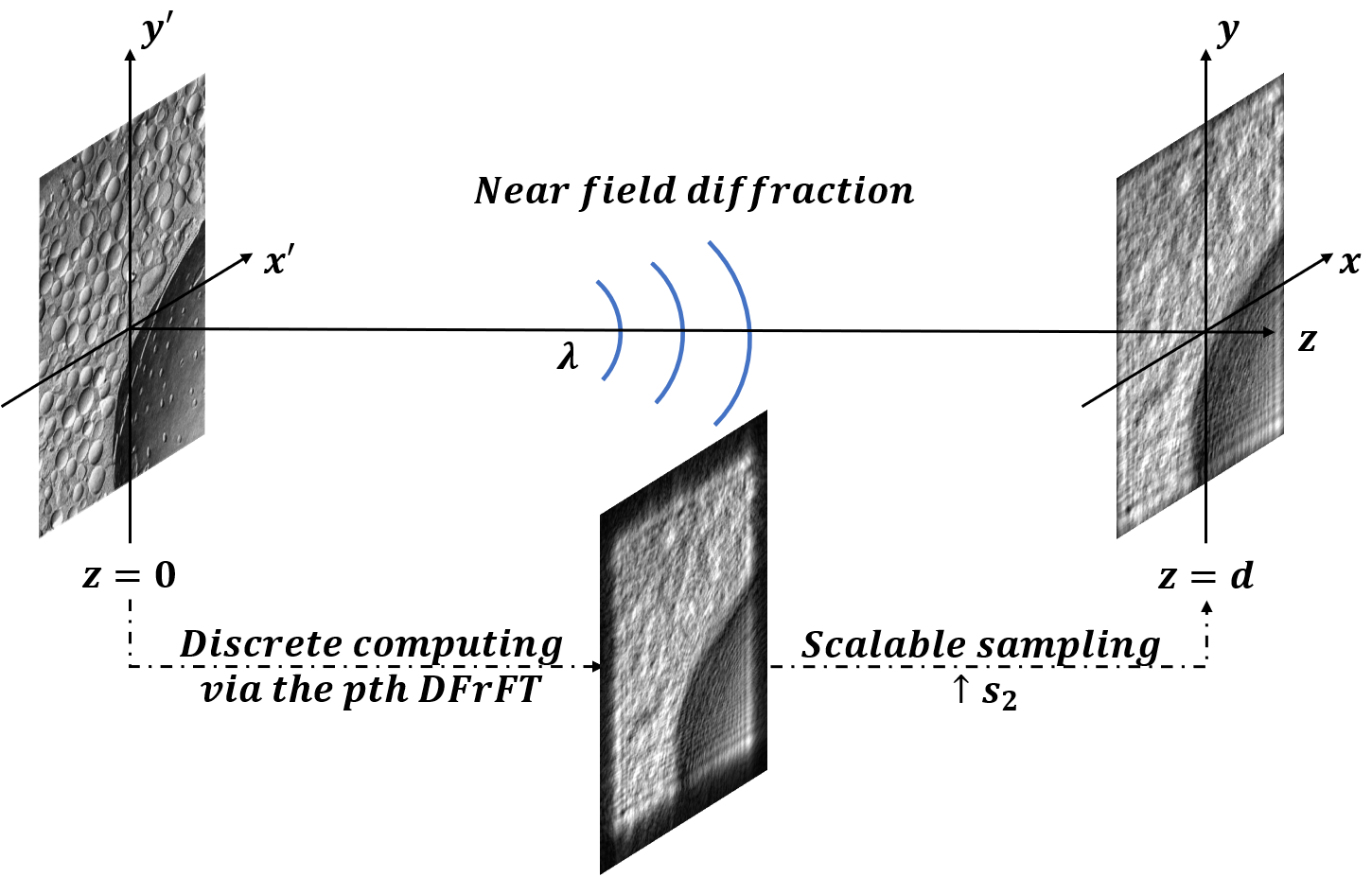}
        \caption{The illustration of FrFT-based Measurement Model. Specifically, the numerical calculation of near-field diffraction can be implemented through the $p$-th discrete fractional Fourier Transform (DFrFT) and the scalable sampling $\uparrow s_2$.}
        \label{fig:frftmeas}
    \end{figure}

    Now, we consider introducing the two-dimensional FrFT of the input aperture $U_0$, which can be defined as
    \begin{equation}
        \mathcal{F}^{p}[U_0](x,y) = \iint K_{\alpha}(x,x') K_{\alpha}(y,y') U_0(x',y')dx'dy',
        \label{eq:frft2d}
    \end{equation}
    where $K_{\alpha}(x,x')$ and $K_{\alpha}(y,y')$ are transform kernels on the x-axis and y-axis, respectively.
    
    To connect the Fresnel integral \eqref{eq:fresnel} and FrFT \eqref{eq:frft2d}, we introduce the scaled fields $\hat{U}_d(x,y) \equiv U_d(s_2 x,s_2 y), \hat{U}_0(x',y') \equiv U_0(s_1 x',s_1 y')$ with $s_1 = \sqrt{\frac{\lambda d}{\mathrm{tan}\alpha}}, s_2 = \sqrt{\frac{\lambda d}{\mathrm{sin}\alpha\mathrm{cos}\alpha}}$. 
    
    Then we can obtain the scaled diffraction field
    \begin{align}
        & \hat{U}_d(x,y) = \frac{e^{i\frac{2\pi}{\lambda} d}}{i\lambda d} \iint \hat{U}_0(x',y')e^{i\frac{\pi}{\lambda d}[(x-x')^2+(y-y')^2]}dx'dy',\\
        & = \frac{e^{i\frac{2\pi}{\lambda} d}}{i\mathrm{tan}\alpha}\iint \hat{U}_0(x',y')e^{i \pi[\frac{x^2+y^2}{\mathrm{sin}\alpha\mathrm{cos}\alpha}+\frac{x'^2+y'^2}{\mathrm{tan}\alpha}-\frac{2(xx'+yy')}{\mathrm{sin}\alpha}]}dx'dy',\\
        & = \frac{e^{i\frac{2\pi}{\lambda} d}}{i\mathrm{tan}\alpha + 1} e^{i\pi \mathrm{tan}\alpha (x^2+y^2)} \mathcal{F}^{p}[\hat{U}_0](x,y).
        \label{eq:scalefrft}
    \end{align}
    
    Considering the intensity-only measurement, the spherical phase factor of \eqref{eq:scalefrft} disappears and we can have
    \begin{equation}
        |\hat{U}_d(x,y)| = \frac{1}{\sqrt{\mathrm{tan}^2\alpha + 1}} |\mathcal{F}^{p}[\hat{U}_0](x,y)|.\\
        \label{eq:frft_meas}
    \end{equation}
	
    Thereby, we conclude that the scaled amplitude distribution of the near-field diffraction can be interpreted as the continuous FrFT magnitude. Given the scale factor $s_1$ and the physical propagation distance $d$, we can know exactly the other scale factor $s_2 = s_1 \sqrt{1+(\lambda d)^2/s_1^4}$ and the corresponding fractional order $p = 2/\pi \times \mathrm{arctan}(\lambda d/s_1^2)$. It can be seen that when the propagation distance $d$ increases, the corresponding fractional order $p$ gradually becomes larger with the range is [0,1]. This is consistent with the Fourier transform corresponding to propagation into the far field.
    
    \textbf{Discrete Computing.} In practical scenarios, the acquisition of the diffraction field is inherently limited to sampling. Therefore, accurate discretization of both the input field and the transform kernel is crucial for numerical calculations to avoid aliasing artifacts. Benefiting from previous studies, various types of discrete fractional Fourier transform (DFrFT) have been developed with distinct strategies and properties \cite{su2019analysis} that could be applied here.
    Prominent examples include the eigenvector decomposition-type DFrFT (ED-DFrFT) \cite{839980}, the improved sampling-type DFrFT (IP-DFrFT) \cite{ozaktas1996digital}, and the closed-form sampling-type DFrFT (CF-DFrFT) \cite{pei2000closed}. Specifically, ED-DFrFT offers orthogonality, additivity, and reversibility properties at the expense of high computational complexity while IP-DFrFT enjoys high discrete accuracy and can be implemented efficiently with lacking unitarity. CF-DFrFT achieves reversible properties by carefully considering sampling interval limits and involves low-complexity calculations in $O(NlogN)$ time. Nevertheless, it is essentially similar to SFT-Fresnel and also faces the problem of numerical errors.
    Therefore, we select the IP-DFrFT approach in this work and present some details as follows.

    Specifically, the samples of the transformed function in \eqref{eq:frft_meas} spaced at the interval $\triangle x$ and $\triangle y$ are obtained as
    \begin{equation}
        |\hat{U}_d(m \triangle x,n \triangle y)| = \frac{1}{\sqrt{\mathrm{tan}^2\alpha + 1}} |\mathcal{F}^{p}[\hat{U}_0](m \triangle x,n \triangle y)|,\\
        \label{eq:discrete_meas}
    \end{equation}
    where $m,n$ goes from $-N/2$ to $N/2$, and $N$ is the sampling number.


    Following the computation scheme of IP-DFrFT, we can get
    \begin{equation}
    \begin{split}
        & | \mathcal{F}^{p}[\hat{U}_0](m \triangle x,n \triangle y)| = A_\alpha^2 |\sum\limits_{m'} \sum\limits_{n'} \hat{U}_0(m' \triangle x', n' \triangle y')\\
        & \times \triangle x' \triangle y' e^{i \pi \mathrm{csc}\alpha [(m-m')^2\triangle x \triangle x' + (n-n')^2\triangle y \triangle y']}\\
        & \times e^{i\pi (\mathrm{cot}\alpha - \mathrm{csc}\alpha)(m'^2\triangle x'^2 + n'^2\triangle y'^2)}|,
    \end{split}
    \label{eq:ip-dfrft}
    \end{equation}
    where $\triangle x' = \triangle x, \triangle y' = \triangle y$ is the spacing interval and $m',n'$ represents the discrete grid in the source field.
    
    Its calculation can be recognized that the discrete source field is first modulated with a chirp function $e^{i \pi (\mathrm{cot}\alpha - \mathrm{csc}\alpha)(m'^2\triangle x'^2 + n'^2\triangle y'^2)}$ and then convoluted by another chirp function $e^{i \pi \mathrm{csc}\alpha [(m'\triangle x')^2 + (n'\triangle y')^2]}$ which can be efficiently implemented by FFT. Note that it can avoid the aliasing error due to the rapid oscillations of the kernel by exploiting the periodicity and additivity of the continuous FrFT\footnote{Note that there is an assumption as $0.5 \leq |p| \leq 1.5$. Taking advantage of the additivity property of FrFT, we can extend the range of parameter $p$ to cover all its values. For example, for the range $0<p<0.5$, we have that $\mathcal{F}^{p} = \mathcal{F}^{p-1+1} = \mathcal{F}^{p-1}\mathcal{F}^{1}$.}.
    Given this, we can get an efficient, accurate, and unified method to calculate the whole diffraction field. It is worth mentioning that it is necessary to perform the dimensional normalization on the fractional order $p$ and the scale factor $s_2$ during the numerical calculations, in order to eliminate the sampling-related influences:
    \begin{equation}
    p = \frac{2}{\pi}\mathrm{arctan}(\frac{\lambda d}{s_1^2} \times \frac{N}{L^2}), s_2 = s_1 \sqrt{1+\frac{(\lambda d \times \frac{N}{L^2})^2}{s_1^4}},\\
    \label{eq:norm}
    \end{equation}
    where $L$ is the length of the input aperture.

    \begin{table}[!t]
        \centering
        \caption{A summary of commonly used Fresnel propagation models as compared to the proposed FrFT measurement model.}
        \label{tab:ovw_nf}
        \setlength{\tabcolsep}{1mm}
        \renewcommand\arraystretch{1.5}
        \begin{tabular}{cccc}		\hline  
            \centering 
            {Method} & {Complexity} & {Pixel Pitch} & {Range} \\  
            \hline
            \centering   
            {SFT-FR} & {1 FFT} & {$\triangle _x = \frac{\lambda d}{N\triangle _x'}$} & {Long-distance} \\ 
            {Fresnel-TF} & {2 FFTs} & {$\triangle _x = \triangle _x'$} & {Short-distance} \\
            {FrFT (this work)} & {2 FFTs} & {$\triangle _x = \sqrt{1+(\frac{\lambda d}{N \triangle _x'^2})^2} \triangle _x'$} & {All-distance} \\
            \hline 
        \end{tabular}
    \end{table}

    \textbf{Scalable Sampling.} Considering the scaling operator introduced in \eqref{eq:frft_meas}, it can be addressed by obtaining the rescaling samples in practice. Specifically, the scale factor $s_1$ can usually be set to 1 in order to be consistent with the real input field, \ie $\hat{U}_0 = U_0$.
    Thereby, the original near-field measurement can be obtained by 
    \begin{equation}
        |U_d(x,y)| = \frac{1}{\sqrt{\mathrm{tan}^2\alpha + 1}} |\mathcal{F}^{p}[U_0](x/s_2,y/s_2)|,\\
        \label{eq:frft_zoom}
    \end{equation}
    where $s_2 = \sqrt{1+(\lambda d N/L^2)^2}$ can be seen as a sampling rate conversion factor.

    Consistent with previous considerations, the sampling interval within the FrFT domain remains $\triangle x'$, consequently leading to the pixel pitch of $s_2 \triangle x'$ within the real observation plane. This, in turn, enables us to directly acquire the intensity measurements of the FrFT through near-field diffraction. To clarify, Table \ref{tab:ovw_nf} shows a summary of the methods mentioned within this work, encompassing pertinent details such as the computational complexity, pixel pitch, and the respective suitable range.
    In addition, we can also adopt digital computation in the FrFT domain to eliminate the scaling operator. 
    It can be noticed that the scaling factor $s_2$ gradually becomes larger as the propagation distance $d$ increases and is always greater than 1, which corresponds to the upsampling case. According to the digital sampling rate conversion, we can implement it through interpolation, low-pass filtering, and decimation in sequence, achieving flexible sampling intervals.
    Overall, the proposed FrFT-based measurement model is illustrated in Fig. \ref{fig:frftmeas}.


    \begin{figure*}[!t]
        \centering
        \includegraphics[width=0.9\linewidth,clip,keepaspectratio]{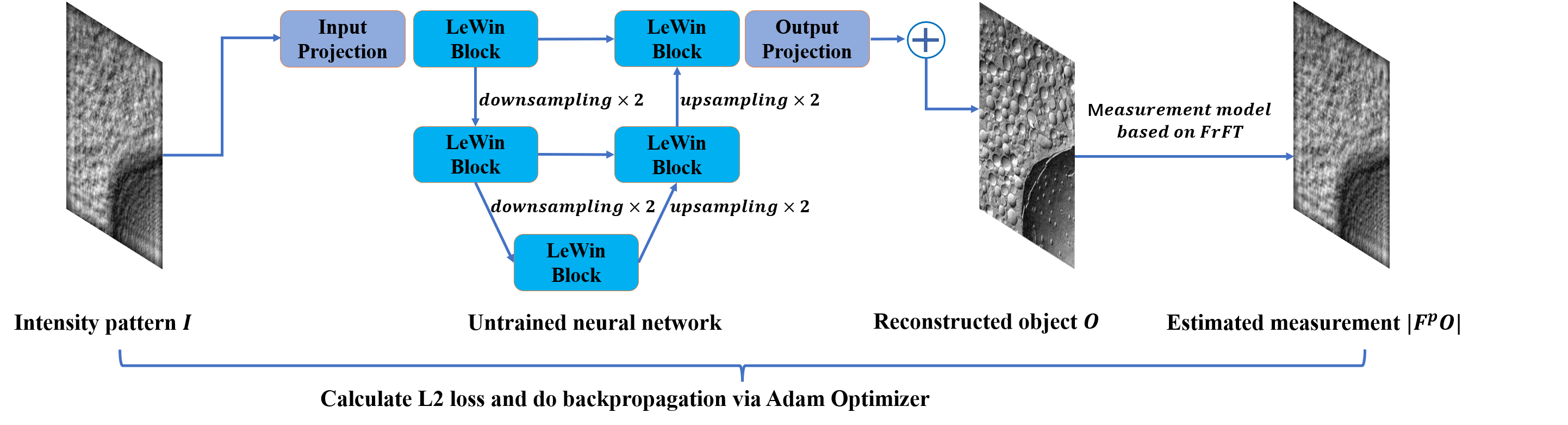}
        \caption{The schematic illustration of SFrFPR. The input to the neural network is a diffraction pattern of an amplitude or phase object, captured in a single snapshot. The neural network processes this input and generates an estimated object. Subsequently, the estimated object is numerically propagated to simulate the resulting diffraction pattern using the proposed FrFT measurement model. To guide the training process, we calculate the mean square error (MSE) between the real and estimated measurements. This MSE serves as the loss value, updating the parameters of the neural network via the auto-differentiation technique.}
        \label{fig:overiew}
    \end{figure*} 

    \subsection{Self-supervised Reconstruction Approach based on UNN}
    This surge in research activity over the last decade has focused on developing reconstruction algorithms for Fourier PR, including iterative optimization approaches \cite{candes2015phase,wang2018phase,metzler2018prdeep,wei2022tfpnp} and neural network approaches \cite{yang2022dynamic,chen2022unsupervised,cha2021deepphasecut,wang2020phase}. In this part, we define the SFrFPR problem for the first time and propose an untrained neural network (UNN) based reconstruction approach for SFrFPR.
    
    \subsubsection{Problem Formulation}
    Based on the proposed FrFT-based measurement model, the SFrFPR problem can be mathematically stated as
    \begin{equation}
        Given \quad I = |\mathcal{F}^{p} O|, \qquad find \quad O,
        \label{eq:FrFPR}
    \end{equation}
    where $I$, $\mathcal{F}^{p}$, and $O$ represent the intensity of the diffraction pattern, the corresponding $p$th-order FrFT measurement model, and the underlying object, respectively.

    Typically, the inverse problem of SFrFPR can be solved within a regularized optimization framework by minimizing the following cost functional:
    \begin{equation}
	\mathop{\mathrm{\mathop{minimize}}}_O \quad \frac{1}{2}||I - |\mathcal{F}^{p} O| ||^2_2 + \beta \mathcal{R}(O),
	\label{eq:optimization}
    \end{equation}
    where $\mathcal{R}(O)$ indicates the regularization term associated with the prior knowledge of objects, and $\beta$ is a parameter that controls the weight of the regularizer. 

    This is a non-convex and non-linear ill-posed problem, mainly caused by the loss of phase. Existing iterative optimization PR methods such as Wirtinger gradient descent \cite{candes2015phase} and plug-and-play methods \cite{metzler2018prdeep,10095976}, are expected to provide transferable ideas to solve this problem.
    However, these methods all rely on accurate forward and backward projections, resulting in the use of ED-DFrFT. The high computational complexity prevents its application to large-scale and real-time reconstruction. The fast discrete FrFT algorithm represented by IP-DFrFT can solve this dilemma, but it lacks unitarity and suffers from numerical errors during the inverse transformation \cite{su2019analysis}, making it unsuitable for these iterative methods.

    \subsubsection{UNN-based Method}
    To address this issue, we put forward an alternative solution, leveraging the concept of the deep image prior (DIP) \cite{ulyanov2018deep} and employing an untrained neural network (UNN) for SFrFPR. The key idea is designing a neural network $f_{NN}(I,\Theta)$ to directly perform the inverse mapping from captured intensity measurement $I$ to the desired object $O$ by adjusting the network's weight $\Theta$ based upon the following empirical risk:

    \begin{align}
        \mathop{\mathrm{\mathop{minimize}}}_\Theta \quad \frac{1}{2}||I - |\mathcal{F}^{p}f_{NN}(I,\Theta)| ||^2_2.
    \label{eq:unn}
    \end{align}

    Compared with \eqref{eq:optimization}, the main difference is that we optimize the neural network's parameters instead of the estimated object directly. This direct problem transformation has brought many benefits. First of all, thanks to the development of neural network technology, this optimization process can be well solved by the auto-differentiation technique. In this way, the proposed method solely requires the differentiability of the forward function while avoiding the need for the backward function.
    In addition, the proposed method operates in a self-supervised manner, utilizing the network's weight adjustment to reconstruct the desired amplitude or phase object, guided by the captured intensity measurement and incorporating the FrFT-based measurement model. Therefore, the proposed method does not rely on obtaining a large number of pairs of ground truth data and corresponding observations. 
    
    \textbf{Overall Pipeline.} The overall pipeline of our method is outlined in Fig. \ref{fig:overiew}. The input to the neural network is a diffraction pattern of an amplitude or phase object, captured in a single snapshot. The neural network processes this input and generates an estimated object. Subsequently, the estimated object is numerically propagated to simulate the resulting diffraction pattern using the proposed FrFT-based measurement model. Engaging a loss function computed between the measurement and the estimated diffraction pattern, the parameters of the neural network are adjusted via the auto-differentiation technique. Notably, this training process only involves the FrFT forward function and unlabeled simulated/measured diffraction patterns. After updating, the trained network can perform the direct inversion from a single intensity pattern to the real space object without requiring the iterative process.

    \textbf{Network Implementation.} In UNN, the architecture of the neural network is very important, because it will introduce the neural network structure prior as a regularization term. Witnessing the powerful representation ability of the Transformer network in large models, we explore its potential as an untrained network for SFrFPR. To be specific, we adopt a general U-shaped Transformer architecture, as delineated by \cite{wang2022uformer}, and build a small hierarchical encoder-decoder network with a tiny computational burden. Given a snapshot pattern as the network input, a convolutional layer with LeakyReLU is used to extract low-level features. Subsequently, these features traverse through two encoder phases, each of which encompasses a LeWin Transformer block and one down-sampling layer.
    The LeWin Transformer block captures long-range dependencies via non-overlapping windows instead of global self-attention, resulting in the low computational cost of high-resolution feature maps \cite{wang2022uformer}. Proceeding further, a bottleneck phase characterized by a LeWin Transformer block is appended to culminate the encoding progression. Then, two decoder phases are followed to recover the features, each of which contains an up-sampling layer and a LeWin Transformer block. Finally, a convolutional layer is applied to output the reconstructed object. 

    While Transformer models have undoubtedly demonstrated remarkable performance in numerous supervised learning tasks, their untapped potential as untrained neural networks remains to be further explored. To the best of our knowledge, this is the first attempt to introduce the Transformer-based architecture into an untrained neural network for solving the PR problem. In Section \ref{sec:experiments}, experimental results demonstrate that the Transformer structure priors leverage both local and global dependencies with better reconstruction performance than convolutional neural networks in a self-supervised scheme.

    \subsection{Theoretical Analysis of SFrFPR}
    \label{sec:analysis}	
    Next, we present theoretical analyses from the perspective of fractional Fourier space-frequency representation to clarify the rationality of the proposed SFrFPR compared to Fourier PR.
    Specifically, we take a one-dimensional energy-limited signal $f(x)$ as an example\footnote{The generalization to the two-dimensional spatial signals considered in this paper is obvious.} and introduce the concept of fractional Wigner–Ville distribution \cite{torres2012fractional}, defined as
    \begin{equation}
        \mathcal{W}_\alpha (x,u) = \frac{|\mathrm{csc}\alpha|}{2\pi} \int f^*(x-\frac{\tau}{2})f(x+\frac{\tau}{2})e^{-i\tau(u\mathrm{csc}\alpha-x\mathrm{cot}\alpha)}d\tau,
    \label{eq:fwvd}
    \end{equation}
    where $u$ denotes the fractional frequency, $\alpha$ represents the fractional angle, and $*$ indicates conjugate operation.
    
    In particular, the fractional Wigner–Ville distribution degenerates into the classical Wigner–Ville distribution when $\alpha = \pi/2$, depicted as
    \begin{equation}
        \mathcal{W} (x,w) = \frac{1}{2\pi} \int f^*(x-\frac{\tau}{2})f(x+\frac{\tau}{2})e^{-i\tau w}d\tau,
    \label{eq:wvd}
    \end{equation}
    where $w$ is the frequency.

    It can be seen that the fractional Wigner–Ville distribution determines the signal representation of the joint space $x$ and fractional frequency $u$, The classic Wigner–Ville distribution provides a signal representation that combines space $x$ and frequency $w$. Comparing the two definitions \eqref{eq:fwvd} and \eqref{eq:wvd}, we can further obtain the relationship between the fractional Wigner–Ville distribution and the classic Wigner–Ville distribution, as
    \begin{equation}
        \mathcal{W}_\alpha (x,u) = \frac{|\mathrm{csc}\alpha|}{2\pi} \mathcal{W} (x,u\mathrm{csc}\alpha-x\mathrm{cot}\alpha).
    \end{equation}

    Therefore, the relationship between fractional frequency $u$ and space $x$ and frequency $w$ is
    \begin{equation}
        w = u\mathrm{csc}\alpha-x\mathrm{cot}\alpha.
    \end{equation}

    Further, we can get
    \begin{equation}
        u = w\mathrm{sin}\alpha + x\mathrm{cos}\alpha.
    \end{equation}

    This shows that the variable of the fractional Fourier transform, that is, the fractional frequency, essentially contains space and frequency information that can depict the frequency changes of the signal over space. In view of this, for any finite energy signal, based on its fractional Fourier transform, a new space-frequency representation of the signal can be defined as
    \begin{equation}
        \mathcal{T}_\alpha (x,u) \triangleq \mathcal{F}_\alpha (w\mathrm{sin}\alpha + x\mathrm{cos}\alpha),
        \label{eq:fsfr}
    \end{equation}
    where $\mathcal{T}_\alpha (x,u)$ is called the fractional Fourier space-frequency representation of the signal.

    Based on the above analysis, the intensity observation of the FrFT is equivalent to the modulus value of the space-frequency representation. Given this, the FrFT measurement has many unique and useful properties that the Fourier measurement does not have, which are beneficial to phase retrieval.
    
    On the one hand, the FrFT measurement contains the amplitude information of the signal, owing to the space-frequency coupling characteristics of the FrFT \cite{almeida1994fractional}. In contrast to the pronounced discrepancy between the Fourier plane and the image plane, the FrFT domain exhibits data distributions that are closely related to those found in the spatial domain. Therefore, some changes in signal on the spatial domain also have effects on the FrFT measurement. For example, signals of space-shift and conjugate-inversion will produce the different FrFT measurements according to the spatial shift and reversal property of the FrFT \cite{tao2009fractional}, respectively, shown in Fig.~\ref{fig:trivial_ambiguities}.
    On the other hand, the FrFT measurement also contains the phase information of the signal. According to \eqref{eq:fsfr}, we can obtain the modulus of the fractional Fourier space-frequency representation of the signal. Furthermore, we can observe how the frequency of the signal changes with space, which is exactly the information contained in the phase of the signal. 
    
    The above analysis shows that although we only collect the amplitude spectrum of the fractional Fourier transform and lose its phase spectrum, the amplitude and phase information about the original signal is not lost and is encoded in the FrFT measurement. Therefore, we can achieve signal recovery from a single FrFT measurement through the reconstruction algorithm.
    On the contrary, the magnitude of the Fourier transform completely loses the information about the original signal, making single-shot PR impossible.
    

    \begin{table}[!t]
	\centering
	\caption{Physical parameter configurations used in the numerical diffraction calculations.}
	\label{tab:numericalprop}
        \setlength{\tabcolsep}{4mm}
        \renewcommand\arraystretch{1.2}
	\medskip
	\begin{tabular}{l c}
		\hline  
		\centering 
		\textsc{Parameter} & \textsc{Value}\\
            \hline
		  {Wavelength} & {$\lambda = 500 \,nm$} \\
            {Spatial Window Length} & {$L = 1000 \,um$} \\
            {Total Sampling Number} & {$N = 512$} \\
            {Rectangular Aperture Width} & {$W = 500 \,um$} \\
		\hline 
        \end{tabular}
	
    \end{table}
    
    \section{Numerical Simulations}
    \label{sec:experiments}
	
    In this section, numerical simulations evaluate the proposed method. First, we validate the effectiveness of the proposed FrFT measurement model in simulated optical settings. Then we provide the results of the proposed reconstructing method. Finally, we further introduce the potential of the proposed method for practical applications.

    \begin{figure}[!t]
        \centering
        \includegraphics[width=0.9\linewidth,clip,keepaspectratio]{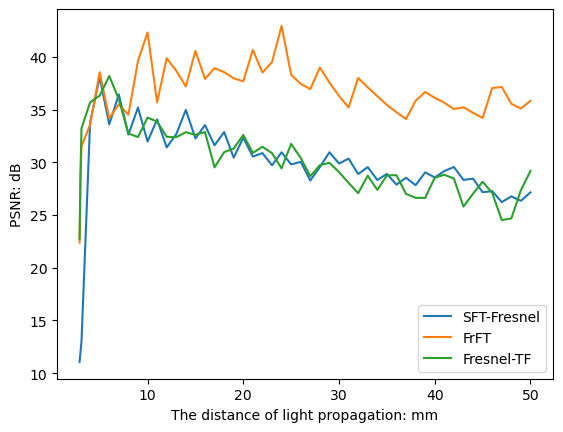}
        \caption{Comparison of the accuracy, evaluated by PSNR, of the SFT-Fresnel, the Fresnel Transfer Function (Fresnel-TF), and the proposed FrFT method in numerical diffraction propagation.}
        \label{fig:numericaldiffraction}
    \end{figure}

    \begin{figure}[t]
        \centering
        \small
        \setlength\tabcolsep{0.8pt}
        \begin{tabular}{cccc}
            {Ground-Truth} & {SFT-Fresnel} & {Fresnel-TF} & {FrFT}\\
            \includegraphics[width=.24\linewidth,clip,keepaspectratio]{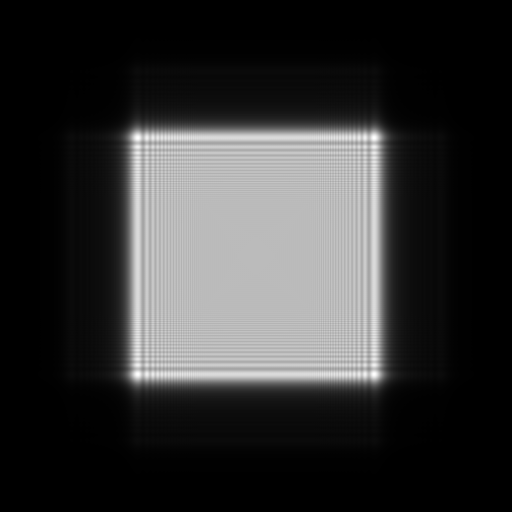} &
            \includegraphics[width=.24\linewidth,clip,keepaspectratio]{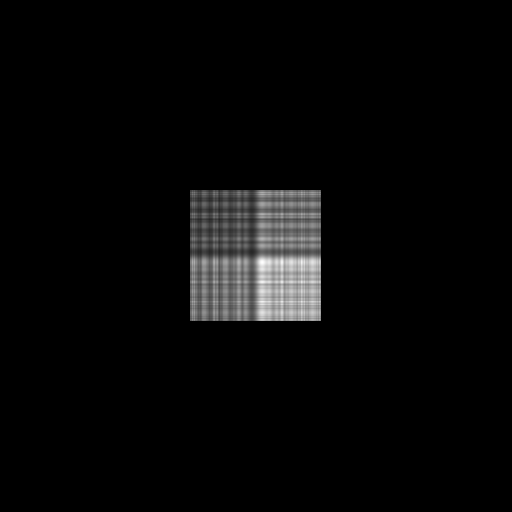} &
            \includegraphics[width=.24\linewidth,clip,keepaspectratio]{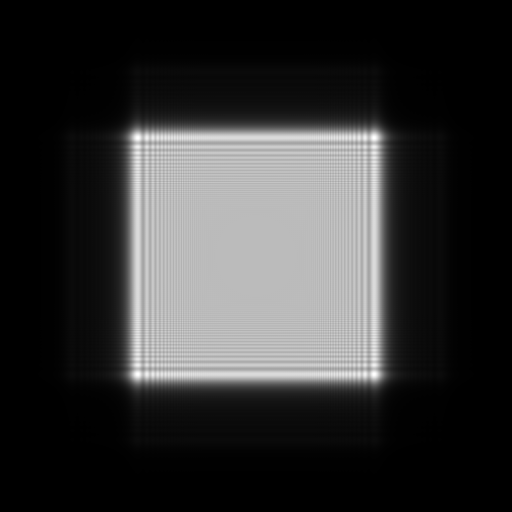} &
            \includegraphics[width=.24\linewidth,clip,keepaspectratio]{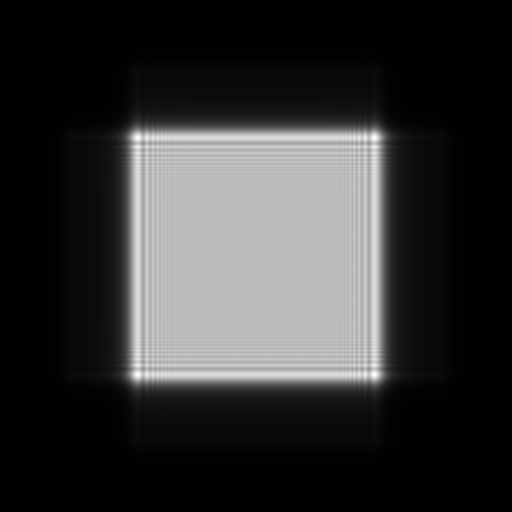} \\
            \includegraphics[width=.24\linewidth,clip,keepaspectratio]{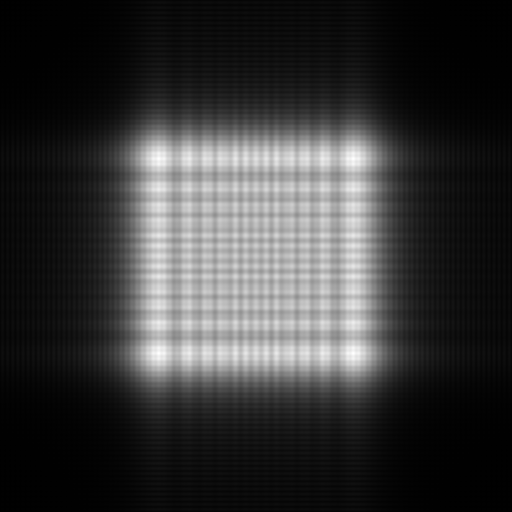} &
            \includegraphics[width=.24\linewidth,clip,keepaspectratio]{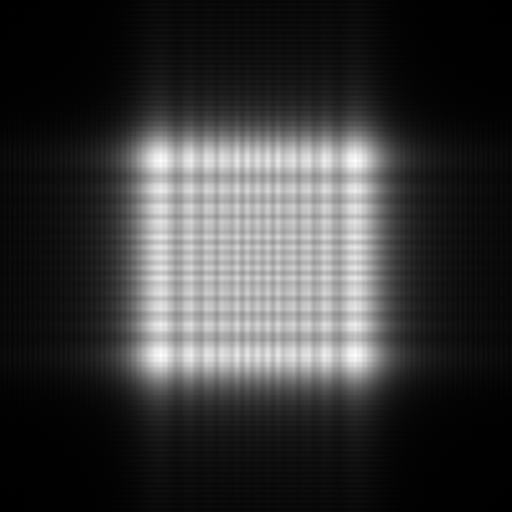} &
            \includegraphics[width=.24\linewidth,clip,keepaspectratio]{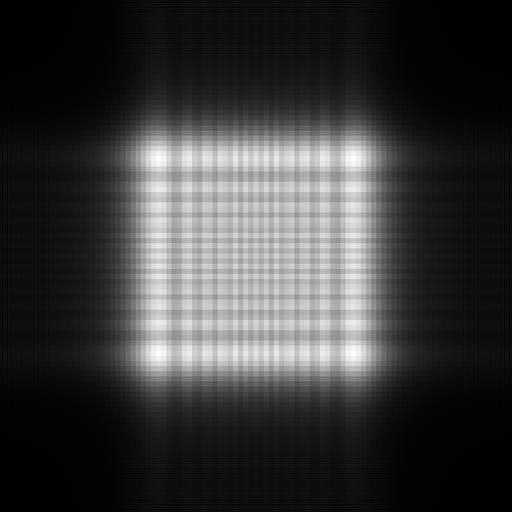} &
            \includegraphics[width=.24\linewidth,clip,keepaspectratio]{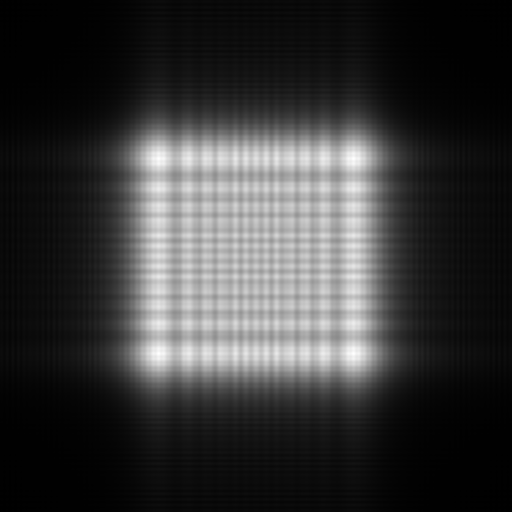} \\
            \includegraphics[width=.24\linewidth,clip,keepaspectratio]{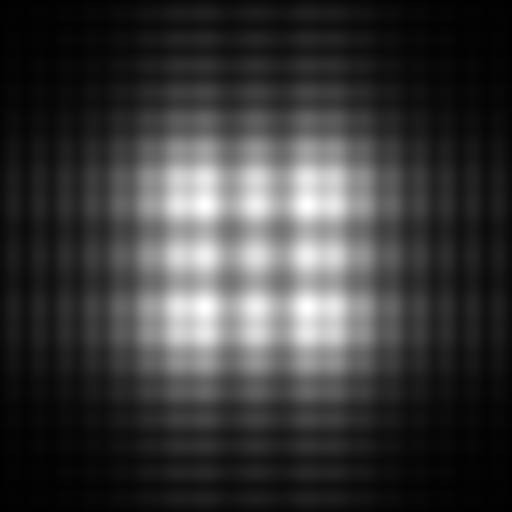} &
            \includegraphics[width=.24\linewidth,clip,keepaspectratio]{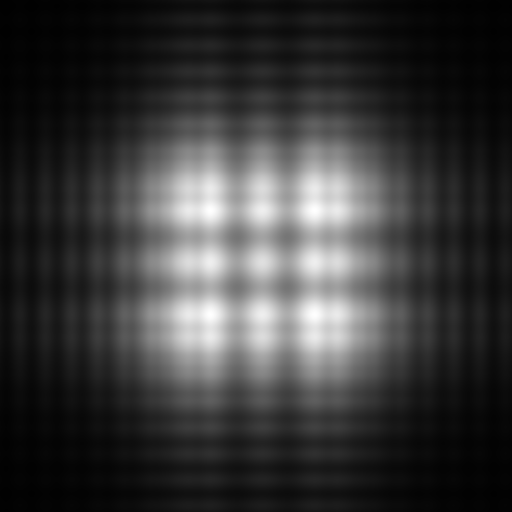} &
            \includegraphics[width=.24\linewidth,clip,keepaspectratio]{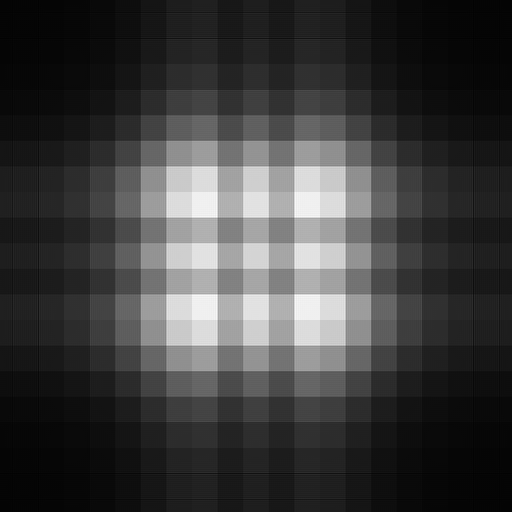} &
            \includegraphics[width=.24\linewidth,clip,keepaspectratio]{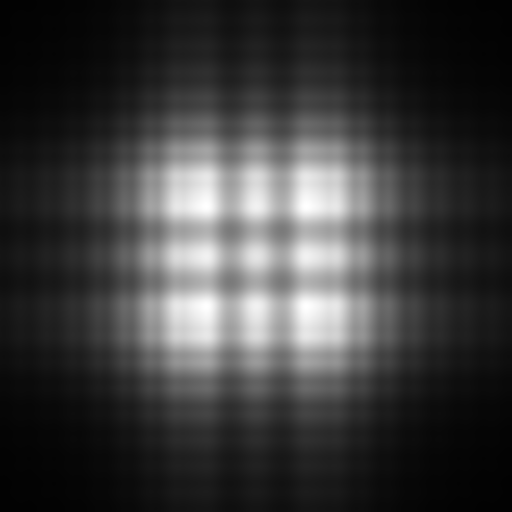} \\
        \end{tabular}
        \caption{Diffraction intensity patterns with the propagation distance ($d=1 \,mm$, $d=10 \,mm$, $d=50 \,mm$ from top to bottom), calculated by the numerical integration (Ground-Truth), single Fourier-transform-based Fresnel model (SFT-Fresnel), Fresnel transfer function model (Fresnel-TF), and the proposed FrFT-based measurement model (FrFT).}
        \label{fig:nearfield_visual}
    \end{figure}	

    \begin{table*}[htbp]
        \centering
        \caption{Average PSNR/SSIM performance comparisons of various reconstruction methods for ``amplitude'' and ``phase'' objects with different fractional Fourier orders on Set12 and Cell8. The best results are labeled in \textbf{bold} and the second are \underline{underlined}.}
        \label{tab:unnbenchmark}
        \renewcommand\arraystretch{1.2}
        \setlength{\tabcolsep}{3mm}{
        \begin{tabular}{ccccccccc}
            \hline  
            \centering 
            \multirow{2}{*}{\textsc{Datasets}} & \multirow{2}{*}{\textsc{Method}} & \multirow{2}{*}{\textsc{Type}} & \multicolumn{5}{c}{\textsc{FrFT Measurement}} & {\textsc{Fourier}}\\\cline{4-8}
            ~ & ~ & ~ & {$p=0.2$} & {$p=0.4$} & {$p=0.5$} & {$p=0.6$} & {$p=0.8$} & {$p=1$}\\ 
            \hline
            \centering
            \multirow{12}{*}{Set12} & {WF} & \multirow{6}{*}{``amplitude''} & {5.92/0.03} & {5.87/0.04} & {22.15/0.72} & {22.09/0.70} & {20.58/0.58} & {10.99/0.11}\\
            ~ & {GAP-tv} & ~ & {6.22/0.01} & {9.04/0.15} & {24.13/0.78} & {23.26/0.76} & {20.42/0.65} & \underline{11.21}/{0.12}\\
            ~ & {prDeep} & ~ & {6.60/0.20} & {6.07/0.16} & {26.59/0.85} & {26.88/0.85} & {27.57/0.85} & {9.37/0.27}\\
            ~ & {PhysenNet} & ~ & {29.40/0.92} & {28.03/0.92} & \underline{29.88/0.93} & \underline{29.96}/{0.89} & {27.58/0.86} & {10.95/0.24}\\
            ~ & {DeepMMSE} & ~ & \underline{29.51/0.92} & \underline{29.67/0.92} & {29.85/0.92} & {29.65}/\underline{0.92} & \underline{29.27}/\textbf{0.92} & {10.36}/\textbf{0.30}\\
            ~ & {Ours} & ~ & \textbf{39.28/0.99} & \textbf{37.75/0.98} & \textbf{36.93/0.98} & \textbf{35.81/0.97} & \textbf{29.98}/\underline{0.90} & \textbf{11.52}/\underline{0.29}\\
            \cline{2-9}
            ~ & {WF} & \multirow{6}{*}{``phase''} & {9.40/0.03} & {12.64/0.26} & {14.90/0.63} & {15.23/0.62} & {12.56/0.50} & \textbf{12.10}/{0.19}\\
            ~ & {GAP-tv} & ~ & {8.50/0.02} & {8.49/0.13} & {12.73/0.56} & {12.85/0.51} & {8.94/0.35} & {10.77/0.27}\\
            ~ & {PhysenNet} & ~ & {14.73/0.76} & \underline{20.74/0.91} & \underline{21.02/0.92} & {20.20/0.85} & \underline{21.96/0.90} & {5.83/0.12}\\
            ~ & {DeepMMSE} & ~ & \underline{18.38/0.86} & {19.01/0.87} & {20.00/0.88} & \underline{21.02/0.88} & {21.90/0.86} & {9.56}/\underline{0.28}\\
            ~ & {Ours} & ~ & \textbf{29.48/0.98} & \textbf{30.57/0.98} & \textbf{30.50/0.98} & \textbf{31.40/0.98} & \textbf{27.93/0.93} & \underline{10.96}/\textbf{0.47}\\
            \cline{2-9}
            \hline
            \centering
            \multirow{12}{*}{Cell8} & {WF} & \multirow{6}{*}{``amplitude''} & {7.03/0.02} & {6.96/0.02} & {21.45/0.76} & {21.06/0.73} & {17.92/0.57} & {11.56/0.05}\\
            ~ & {GAP-tv} & ~ & {7.34/0.02} & {10.26/0.15} & {20.78/0.73} & {19.85/0.69} & {16.97/0.55} & \textbf{12.05}/{0.06}\\
            ~ & {prDeep} & ~ & {8.41/0.09} & {8.62/0.10} & {23.93/0.77} & {23.16/0.77} & {21.53/0.71} & {7.39/0.06}\\
            ~ & {PhysenNet} & ~ & \underline{25.21/0.85} & \underline{28.12/0.93} & \underline{29.54/0.94} & \underline{29.30}/\textbf{0.94} & {22.65/0.74} & {11.71/0.15}\\
            ~ & {DeepMMSE} & ~ & {25.21/0.83} & {25.31/0.83} & {25.49/0.84} & {25.48/0.84} & \underline{24.45/0.81} & {11.70}/\textbf{0.20}\\
            ~ & {Ours} & ~ & \textbf{37.57/0.99} & \textbf{35.41/0.98} & \textbf{33.66/0.97} & \textbf{30.64}/\underline{0.93} & \textbf{24.78/0.81} & \underline{11.90/0.17}\\
            \cline{2-9}
            ~ & {WF} & \multirow{6}{*}{``phase''} & {8.76/0.03} & {11.65/0.16} & {15.53/0.65} & {15.93/0.64} & {11.54/0.36} & {11.70/0.09}\\
            ~ & {GAP-tv} & ~ & {8.65/0.02} & {8.68/0.08} & {14.31/0.48} & {14.36/0.46} & {10.83/0.30} & \underline{11.76}/{0.13}\\
            ~ & {PhysenNet} & ~ & \underline{21.39/0.91} & \underline{21.68/0.91} & \underline{22.10/0.91} & \underline{20.61/0.88} & {17.50/\underline{0.74}} & {7.41/0.08}\\
            ~ & {DeepMMSE} & ~ & {16.68/0.70} & {17.68/0.72} & {18.11/0.73} & {17.54/0.73} & {\underline{17.76}/0.72} & {9.66}/\underline{0.15}\\
            ~ & {Ours} & ~ & \textbf{30.69/0.99} & \textbf{32.15/0.98} & \textbf{31.53/0.98} & \textbf{29.47/0.96} & \textbf{22.85/0.84} & \textbf{12.30/0.28}\\
            \hline
        \end{tabular}}
        
    \end{table*}
    
    \begin{figure*}[!t]
        \centering
        \setlength\tabcolsep{0.8pt}
        \begin{tabular}{cccccccc}
            {Measurement} & {WF} & {GAP-tv} & {prDeep} & {PhysenNet} & {DeepMMSE} & {Ours} & {Ground-Truth}\\
            \includegraphics[width=.12\linewidth,clip,keepaspectratio]{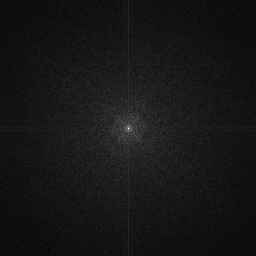} &
            \includegraphics[width=.12\linewidth,clip,keepaspectratio]{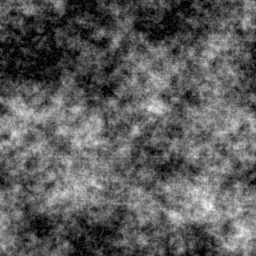} &
            \includegraphics[width=.12\linewidth,clip,keepaspectratio]{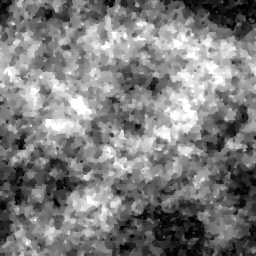} &
            \includegraphics[width=.12\linewidth,clip,keepaspectratio]{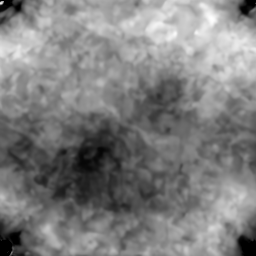} &
            \includegraphics[width=.12\linewidth,clip,keepaspectratio]{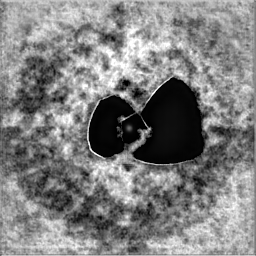} &
            \includegraphics[width=.12\linewidth,clip,keepaspectratio]{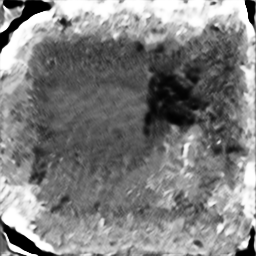} &
            \includegraphics[width=.12\linewidth,clip,keepaspectratio]{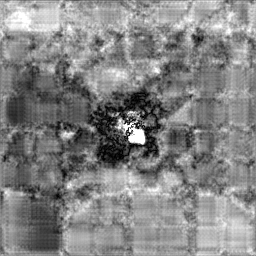} &
            \includegraphics[width=.12\linewidth,clip,keepaspectratio]{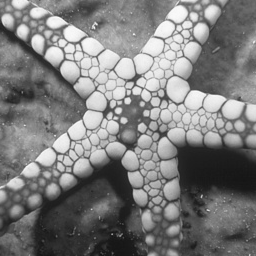} \\
            \footnotesize Fourier ($p = 1$) & \footnotesize 9.64 dB & \footnotesize 10.57 dB & \footnotesize 8.82 dB & \footnotesize 9.16 dB & \footnotesize 8.46 dB & \footnotesize 10.08 dB & \footnotesize PSNR\\
            \includegraphics[width=.12\linewidth,clip,keepaspectratio]{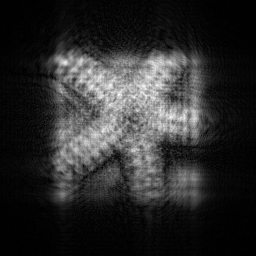} &
            \includegraphics[width=.12\linewidth,clip,keepaspectratio]{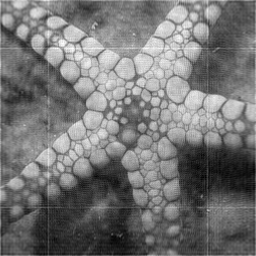} &
            \includegraphics[width=.12\linewidth,clip,keepaspectratio]{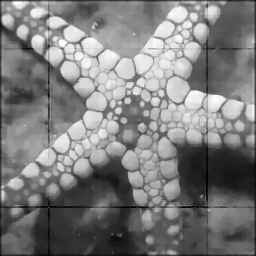} &
            \includegraphics[width=.12\linewidth,clip,keepaspectratio]{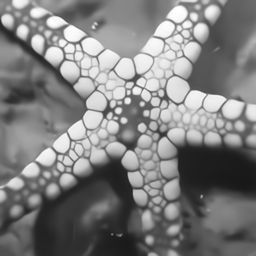} &
            \includegraphics[width=.12\linewidth,clip,keepaspectratio]{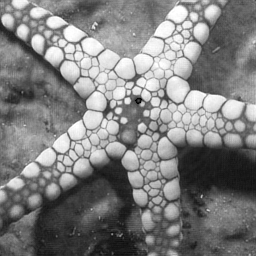} &
            \includegraphics[width=.12\linewidth,clip,keepaspectratio]{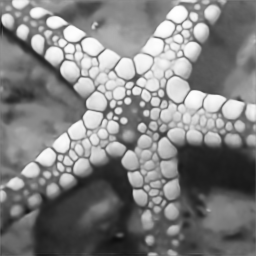} &
            \includegraphics[width=.12\linewidth,clip,keepaspectratio]{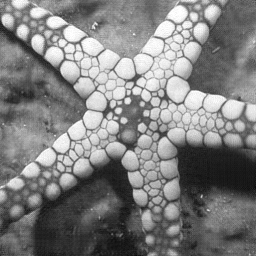} &
            \includegraphics[width=.12\linewidth,clip,keepaspectratio]{figures/Results_amplitude/04.png} \\
            \footnotesize FrFT ($p = 0.6$)  & \footnotesize  22.17 dB & \footnotesize 23.90 dB & \footnotesize 30.06 dB & \footnotesize  32.73 dB & \footnotesize  28.61 dB & \footnotesize  33.79 dB & \footnotesize PSNR\\
            \includegraphics[width=.12\linewidth,clip,keepaspectratio]{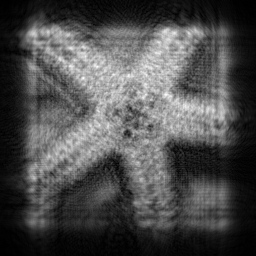} &
            \includegraphics[width=.12\linewidth,clip,keepaspectratio]{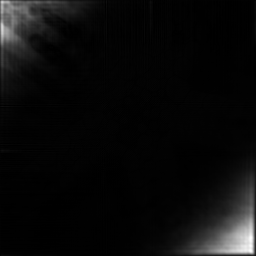} &
            \includegraphics[width=.12\linewidth,clip,keepaspectratio]{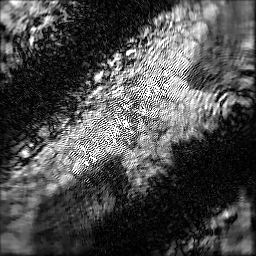} &
            \includegraphics[width=.12\linewidth,clip,keepaspectratio]{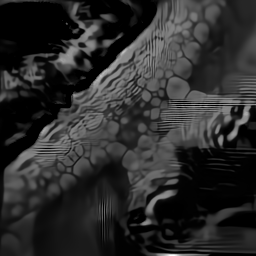} &
            \includegraphics[width=.12\linewidth,clip,keepaspectratio]{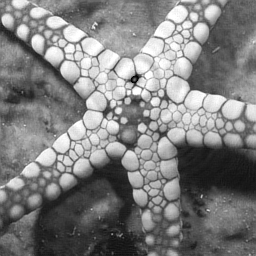} &
            \includegraphics[width=.12\linewidth,clip,keepaspectratio]{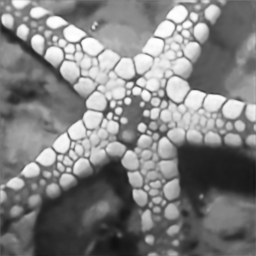} &
            \includegraphics[width=.12\linewidth,clip,keepaspectratio]{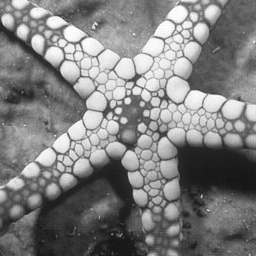} &
            \includegraphics[width=.12\linewidth,clip,keepaspectratio]{figures/Results_amplitude/04.png} \\
            \footnotesize FrFT ($p = 0.4$)  & \footnotesize 5.62 dB & \footnotesize  9.49 dB & \footnotesize  6.16 dB & \footnotesize  30.17 dB & \footnotesize  27.72 dB & \footnotesize  38.42 dB & \footnotesize  PSNR\\
        \end{tabular}
        \caption{Reconstruction results (amplitude objects) of six PR methods on Fourier measurement ($p = 1$) and FrFT measurements with different orders ($p = 0.6$ and $p = 0.4$, respectively) from top to bottom.}
        \label{fig:benchmark_amp}
    \end{figure*}	

    \begin{figure*}[t]
        \centering
        \setlength\tabcolsep{0.8pt}
        \begin{tabular}{ccccccc}
            {Measurement} & {WF} & {GAP-tv} & {PhysenNet} & {DeepMMSE} & {Ours} & {Ground-Truth}\\
            \includegraphics[width=.12\linewidth,clip,keepaspectratio]{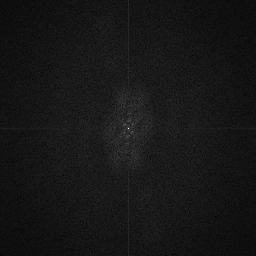} &
            \includegraphics[width=.12\linewidth,clip,keepaspectratio]{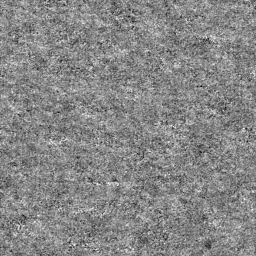} &
            \includegraphics[width=.12\linewidth,clip,keepaspectratio]{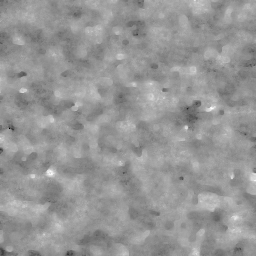} &
            \includegraphics[width=.12\linewidth,clip,keepaspectratio]{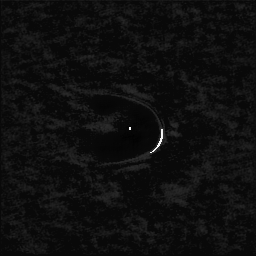} &
            \includegraphics[width=.12\linewidth,clip,keepaspectratio]{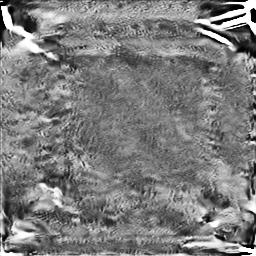} &
            \includegraphics[width=.12\linewidth,clip,keepaspectratio]{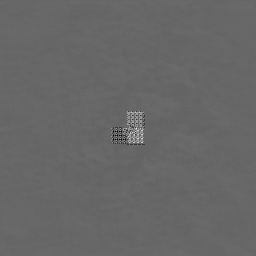} &
            \includegraphics[width=.12\linewidth,clip,keepaspectratio]{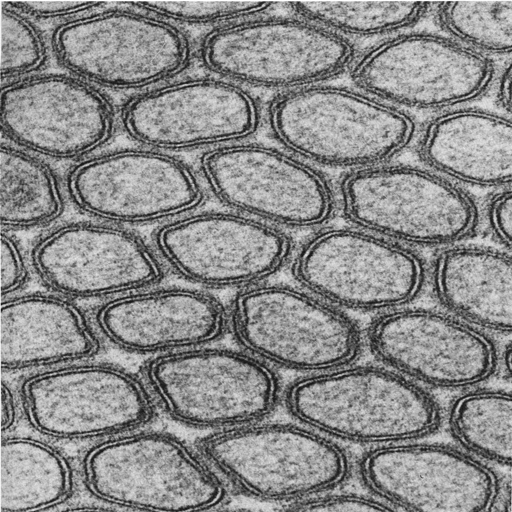} \\
            \footnotesize Fourier ($p = 1$) & \footnotesize 14.28 dB & \footnotesize 13.56 dB & \footnotesize 5.17 dB & \footnotesize 11.78 dB & \footnotesize 12.00 dB & \footnotesize PSNR\\
            \includegraphics[width=.12\linewidth,clip,keepaspectratio]{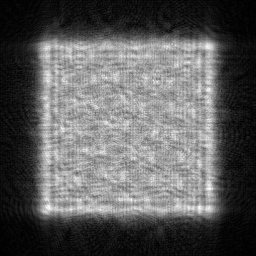} &
            \includegraphics[width=.12\linewidth,clip,keepaspectratio]{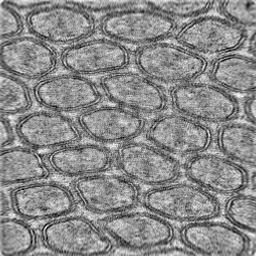} &
            \includegraphics[width=.12\linewidth,clip,keepaspectratio]{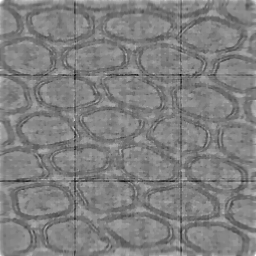} &
            \includegraphics[width=.12\linewidth,clip,keepaspectratio]{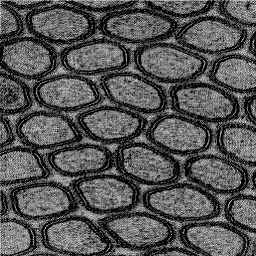} &
            \includegraphics[width=.12\linewidth,clip,keepaspectratio]{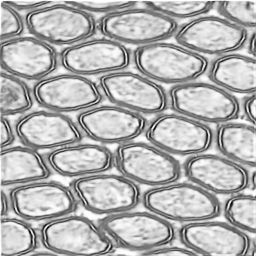} &
            \includegraphics[width=.12\linewidth,clip,keepaspectratio]{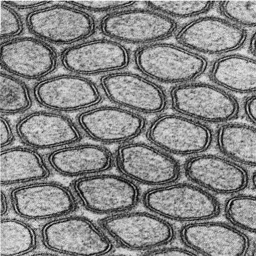} &
            \includegraphics[width=.12\linewidth,clip,keepaspectratio]{figures/Results_phase/16.png} \\
            \footnotesize FrFT ($p = 0.5$)  & \footnotesize  19.59 dB & \footnotesize 16.25 dB & \footnotesize  11.55 dB & \footnotesize  16.81 dB & \footnotesize  25.51 dB & \footnotesize PSNR\\
            \includegraphics[width=.12\linewidth,clip,keepaspectratio]{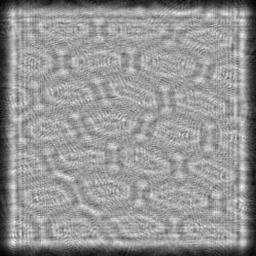} &
            \includegraphics[width=.12\linewidth,clip,keepaspectratio]{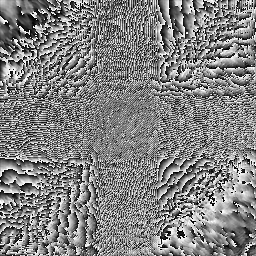} &
            \includegraphics[width=.12\linewidth,clip,keepaspectratio]{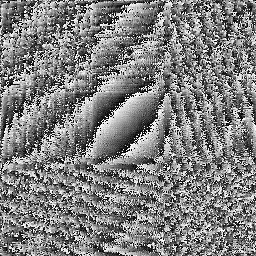} &
            \includegraphics[width=.12\linewidth,clip,keepaspectratio]{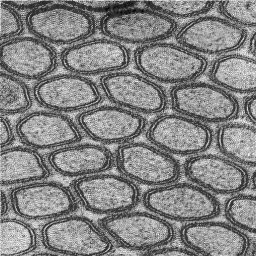} &
            \includegraphics[width=.12\linewidth,clip,keepaspectratio]{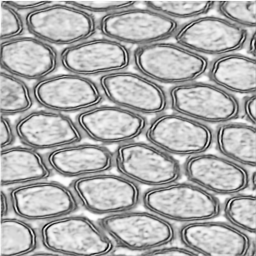} &
            \includegraphics[width=.12\linewidth,clip,keepaspectratio]{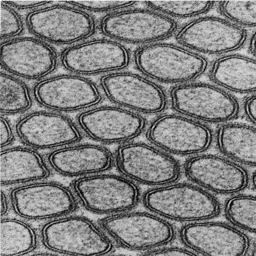} &
            \includegraphics[width=.12\linewidth,clip,keepaspectratio]{figures/Results_phase/16.png} \\
            \footnotesize FrFT ($p = 0.2$)  & \footnotesize 9.78 dB & \footnotesize 9.15 dB & \footnotesize  20.22 dB & \footnotesize  17.72 dB & \footnotesize  21.85 dB & \footnotesize  PSNR\\
        \end{tabular}
        \caption{Reconstruction results (phase objects) of five PR methods on Fourier measurement ($p = 1$) and FrFT measurements with different orders ($p = 0.5$ and $p = 0.2$, respectively) from top to bottom.}
        \label{fig:benchmark_phs}
    \end{figure*}	
   
    \subsection{Evaluation for the FrFT-based measurement model}
    \subsubsection{Implementation details}
    To facilitate a comprehensive assessment of the proposed method, we conducted a computational simulation involving the diffraction propagation of a two-dimensional rectangular aperture under typical physical parameter configurations. Specifically, the spatial length of the sampling windows and the number of samplings are $1000$ um and $512$ in the source and destination plane, respectively. The width of the rectangular aperture and the wavelength are $500$ um and $500$ nm, respectively. The range of propagation distance is $1\sim 50$ mm. The detailed parameter values in numerical calculations are listed in Table \ref{tab:numericalprop}.
    Moreover, the proposed FrFT-based measurement model was realized on Pytorch, thus embracing the modern GPU acceleration.
    
    \subsubsection{Verify the accuracy of the proposed method}
    To verify the accuracy of the proposed FrFT-based measurement model, we conduct two-dimensional diffraction computations using three distinct methods, \ie single Fourier-transform-based Fresnel model (SFT-Fresnel), Fresnel transfer function model (Fresnel-TF), and the proposed FrFT-based measurement model (FrFT). For quantitative analysis, the reference field is provided by numerical integration of the Fresnel integral \eqref{eq:fresnel} via the trapezium rule.
    And we achieved the scalable sampling via post-processing to ensure consistent resolution and pixel pitch size for comparison.

    The accuracy of each method is assessed by comparing the peak-signal-to-noise ratio (PSNR) of the diffraction pattern against the reference, with the results presented as a function of propagation distance in Fig. \ref{fig:numericaldiffraction}. 
    It can be observed that the accuracy of the Fresnel-TF method deteriorates with increasing propagation distance, while the SFT-Fresnel method is unsuitable for numerical propagation over short distances. In contrast, the proposed FrFT method demonstrates remarkable accuracy and features suitability for both short-distance and long-distance propagation.
    An illustrative examination of the diffraction intensity patterns computed through these methods is presented in Figure \ref{fig:nearfield_visual}. Notably, the SFT-Fresnel method manifests pronounced numerical errors within regions corresponding to short propagation distances. 
    Conversely, while the Fresnel-TF method shows good performance when close to the source field, numerical errors become increasingly apparent at longer propagation distances.
    In contrast, the proposed FrFT-based model demonstrates consistency with numerical integration, providing similar results over the entire propagation distance range without apparent aliasing errors in both short- and long-range scenarios.

    \subsection{Evaluation for the UNN-based reconstructing method} 
    
    \begin{figure}[!t]
        \centering
        \subfloat[The curve of measurement loss.]{\includegraphics[width=.9\columnwidth]{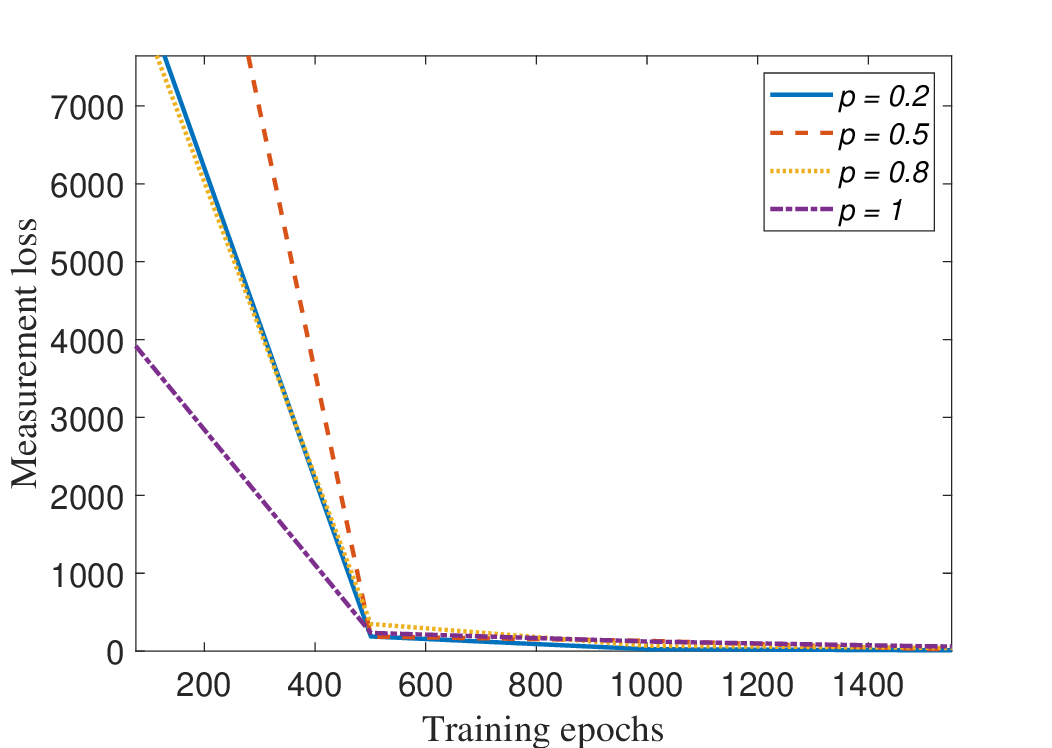}}\vspace{-4.5mm} \\
        \subfloat[The curve of reconstruction quality (PSNR).]{\includegraphics[width=.9\columnwidth]{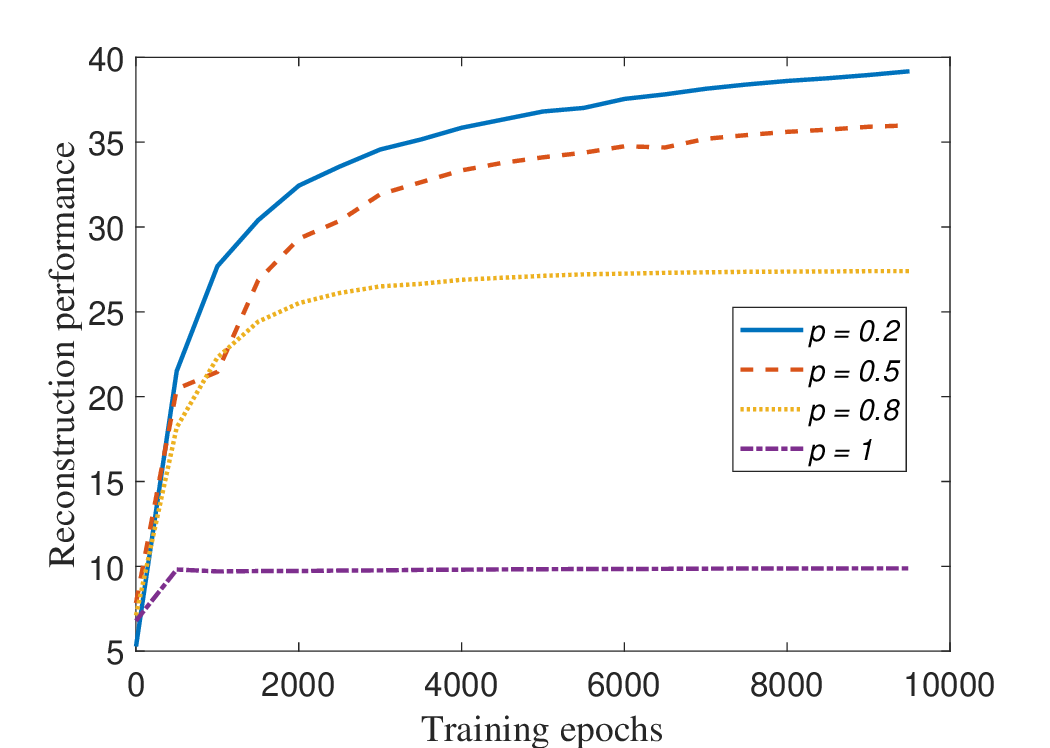}}
        \caption{Convergence behaviors of the FrFT measurements with different orders using the proposed UNN-based method. (a) shows the measurement loss and (b) presents the corresponding reconstruction quality (PSNR) over the training epochs.}
        \label{fig:convergence}
    \end{figure}	

    \begin{figure}[!t]
        \centering
        \small
        \setlength\tabcolsep{0.8pt}
        \begin{tabular}{cccc}
            {Initialization} & {Epoch-100} & {Epoch-200} & {Epoch-500}\\
            \includegraphics[width=.24\linewidth,clip,keepaspectratio]{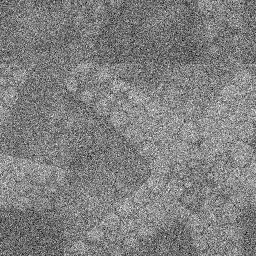} &
            \includegraphics[width=.24\linewidth,clip,keepaspectratio]{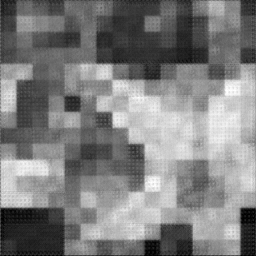} &
            \includegraphics[width=.24\linewidth,clip,keepaspectratio]{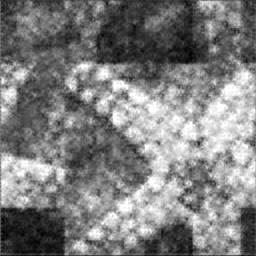} &
            \includegraphics[width=.24\linewidth,clip,keepaspectratio]{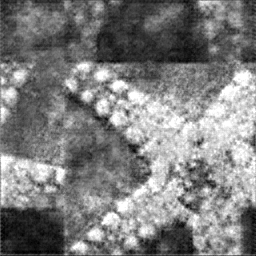} \\
            \footnotesize Fourier ($p = 1$)  & \footnotesize 9.46 dB & \footnotesize 9.21 dB & \footnotesize 9.23 dB\\
            \includegraphics[width=.24\linewidth,clip,keepaspectratio]{figures/init_shift/04_shift64.png} &
            \includegraphics[width=.24\linewidth,clip,keepaspectratio]{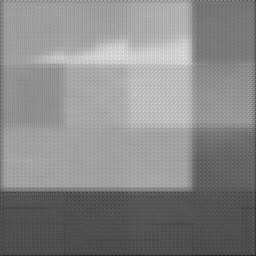} &
            \includegraphics[width=.24\linewidth,clip,keepaspectratio]{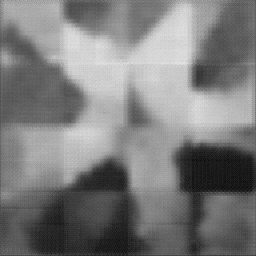} &
            \includegraphics[width=.24\linewidth,clip,keepaspectratio]{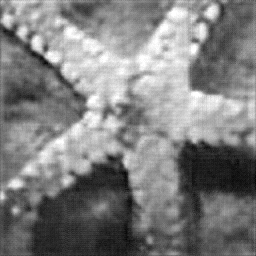} \\
            \footnotesize FrFT ($p = 0.5$) & \footnotesize  14.02 dB & \footnotesize  17.68 dB & \footnotesize  20.40 dB\\
            \includegraphics[width=.24\linewidth,clip,keepaspectratio]{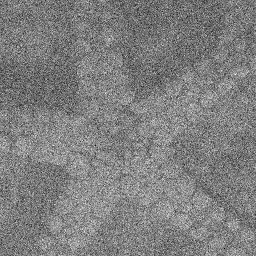} &
            \includegraphics[width=.24\linewidth,clip,keepaspectratio]{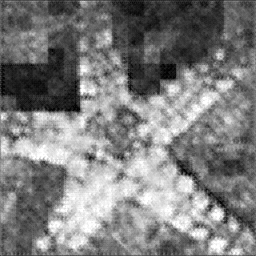} &
            \includegraphics[width=.24\linewidth,clip,keepaspectratio]{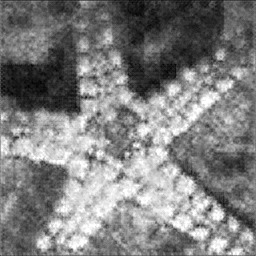} &
            \includegraphics[width=.24\linewidth,clip,keepaspectratio]{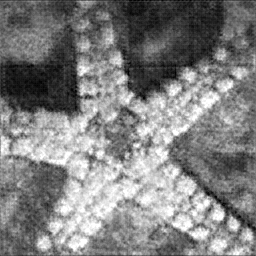} \\
            \footnotesize Fourier ($p = 1$)  & \footnotesize 9.14 dB & \footnotesize 9.09 dB & \footnotesize 9.05 dB\\
            \includegraphics[width=.24\linewidth,clip,keepaspectratio]{figures/init_inversion/04_rotate180.png} &
            \includegraphics[width=.24\linewidth,clip,keepaspectratio]{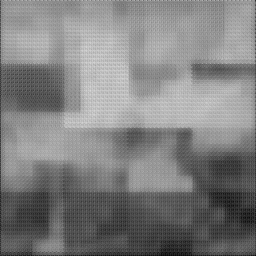} &
            \includegraphics[width=.24\linewidth,clip,keepaspectratio]{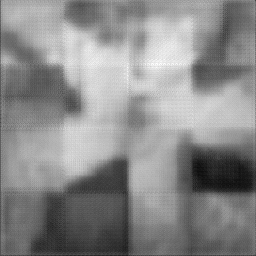} &
            \includegraphics[width=.24\linewidth,clip,keepaspectratio]{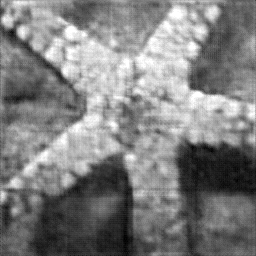} \\
            \footnotesize FrFT ($p = 0.5$) & \footnotesize  14.07 dB & \footnotesize  16.65 dB & \footnotesize  20.56 dB\\
        \end{tabular}
        \caption{The effects of two initializations (the translation and inversion of signal with some Gaussian noise) on the reconstruction process from the Fourier transform measurement ($p = 1$) and the FrFT measurement ($p = 0.5$) using the proposed UNN-based method.}
        \label{fig:init}
    \end{figure}	

    \begin{figure*}[t]
        \centering
        \includegraphics[width=0.9\linewidth,clip,keepaspectratio]{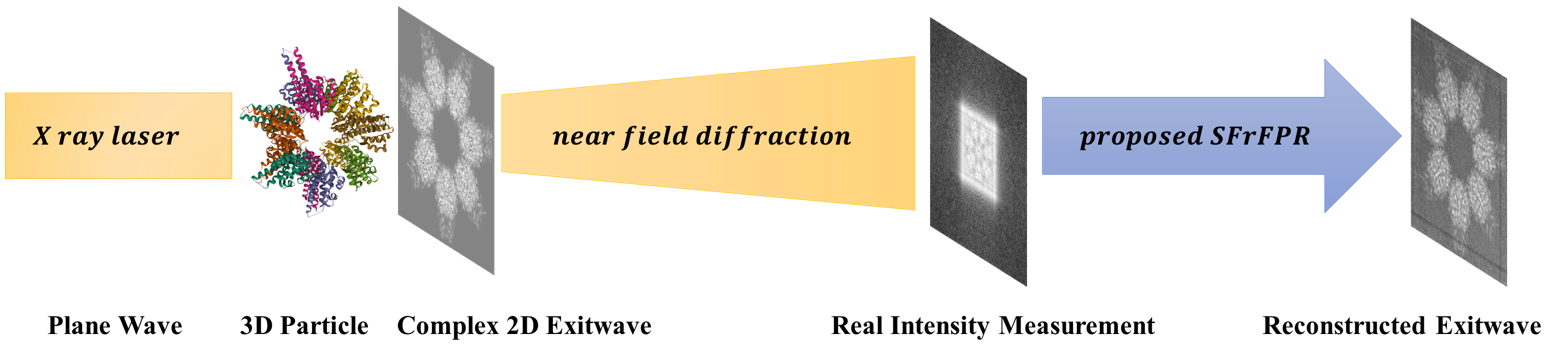}
        \caption{The schematic illustration of support-free CDI. The X-ray laser produces plane waves that illuminate a 3D particle and project it into a 2D exitwave. Through near-field diffraction, a single-shot intensity pattern based on the FrFT measurement is obtained by a detector. And then, a feasible exitwave can be retrieved via the proposed UNN-based reconstruction method.}
        \label{fig:overiew_cdi}
    \end{figure*} 

    \subsubsection{Implementation Details}
    The proposed reconstruction method was implemented based on the PyTorch version $2.0.0$ platform with Python $3.9$ via one Nvidia GeForce GTX 1080 Ti GPU. 
    During the training process, the model was optimized using the Adam optimizer for a total of $10000$ epochs, with the learning rate empirically set at $2e^{-4}$. 
    Notably, our method does not require any external training data apart from the input measurements themselves. To evaluate the effectiveness of the proposed method, we utilized two testing datasets Set12 \cite{zhang2017beyond} and Cell8\footnote{This dataset comprises a collection of eight distinct cell images, namely Brown fat cell, Peritoneal macrophage, Intestinal epithelial cell, Epithelial cell, Blood cell, Auditory hair cell, Acinar cell, and Chromoffin cell. These original cell images are available for download from the website http://www.cellimagelibrary.org/browse/celltype.}. For the sake of unification, all images were resized to a uniform dimension of $256\times 256$.
    
    \subsubsection{Benchmark on SFrFPR} 
    To verify the performance of the proposed method, we mainly compare it against one classic PR approach, namely Wirtinger Flow (WF) \cite{candes2015phase}, two plug-and-play approaches GAP-tv \cite{10095976} and prDeep \cite{metzler2018prdeep}, and two state-of-the-art untrained neural network (UNN) approaches PhysenNet \cite{wang2020phase} and DeepMMSE \cite{chen2022unsupervised}, which are implemented to be extended to solve the SFrFPR problem. Among them, various hyperparameters of these algorithms are set to the optimal. Here we further consider the reconstruction of amplitude-only and phase-only objects from measurements in different fractional Fourier orders\footnote{prDeep is limited to real-valued reconstruction, so phase-only objects are not considered here \cite{metzler2018prdeep}.}. Table~\ref{tab:unnbenchmark} reports the average recovery accuracy in terms of mean peak-signal-to-noise ratio (PSNR) and structure similarity index measure (SSIM) on two testing datasets Set12 and Cell8. It can be observed that optimization-based iterative algorithms such as WF, GAP-tv, and prDeep, can effectively reconstruct objects from some fractional Fourier measurements ($p = 0.5,0.6,0.8$) while all fail to recover images from the Fourier transform measurement ($p = 1$). Unfortunately, due to the inverse transform error of the fast fractional Fourier transform discrete algorithm, these iterative algorithms cannot achieve good results at some specific FrFT orders ($p = 0.2,0.4$).
    On the contrary, all UNN methods consistently perform well on FrFT measurements with different fractional orders except $p = 1$ which denotes the Fourier transform measurement. Moreover, the reconstructing performance can be dramatically improved by the proposed transformer-based UNN when using the FrFT measurement.
    
    For visual comparison, Fig.~\ref{fig:benchmark_amp} and Fig.~\ref{fig:benchmark_phs} present the reconstruction results of different PR methods on an amplitude-only and phase-only object from various FrFT measurements, respectively. It can be seen that all UNN methods can reconstruct the satisfactory amplitude and phase of objects from the single FrFT measurement but fail from the Fourier transform measurement. Note that the proposed method can significantly improve the reconstruction quality. In addition, the classic iterative method WF can also reconstruct the object from a single FrFT measurement with a suitable order, which cannot be achieved in the Fourier case.

    \subsubsection{Algorithmic Investigation}   
    To further investigate the effectiveness of the proposed method, we show the convergence behaviors of the proposed method from different FrFT measurements. Fig.~\ref{fig:convergence} presents that the measurement loss calculated by mean square error (MSE) and reconstruction quality indicated by PSNR varies with training epochs of the proposed method on testing an amplitude object. It can be found that the measurement loss can gradually decrease and converge to a very small value for each FrFT order. However, the proposed method suffers from serious stagnation and produces a poor reconstruction quality (low PSNR) from the Fourier transform measurement. On the contrary, the FrFT measurement can be well-combined with the UNN priors to effectively avoid the stagnation problem and improve the reconstruction quality with increasing PSNR. 

    In order to verify that the proposed FrFT measurement can effectively overcome the ambiguity problem, we use two special but representative initializations to illuminate it. Specifically, we shift and flip the original image adding Gaussian random noise as the input of reconstruction methods. Then we utilize the proposed UNN-based method to reconstruct the original image from its Fourier transform ($p = 1$) and FrFT ($p = 0.5$) measurement, respectively. Fig.~\ref{fig:init} presents the reconstructed results of the intermediate reconstruction process. It can be found that the proposed method can gradually reconstruct a correct object by eliminating the effect of translation or inversion from the FrFT measurement. On the contrary, the reconstruction method suffers from serious stagnation and produces an inaccurate solution from the Fourier transform measurement. While such trivial ambiguities are probably acceptable, they will compete with the correct solution and confuse the reconstructing algorithms in practice. Thus, the removal of ambiguities via the FrFT measurement can greatly alleviate the stagnation problem and improve the reconstruction performance.


    \begin{table*}[!t]
	\centering
	\caption{Summary of numerical optical parameters for X-ray CDI experiments. The energy of the X-ray laser and the information in the source plane are determined, including the field of view, the number of discrete points, and the pixel size. According to different physical propagation distances, we can sequentially obtain corresponding information in the detector plane through the proposed FrFT-based measurement model.}
	\label{tab:exp_params}
        \setlength{\tabcolsep}{5mm}
        \renewcommand\arraystretch{1.5}
	\medskip
	\begin{tabular}{cccc}
		\hline  
		\centering 
		\textsc{Source Plane} & \multicolumn{3}{c}{\textsc{Detector Plane}}\\
            \hline
		  {\makecell{X-ray energy: \\ $5 \,keV$}} & {\makecell{Propagation distance: \\ $d = 0.1 \,m$}} & {\makecell{Propagation distance: \\ $d = 0.25 \,m$}} & {\makecell{Propagation distance: \\ $d = 10 \,m$}}\\
            \multirow{2}{*}{\makecell{Corresponding wavelength: \\ $0.248 \,nm$}} & {\makecell{Corresponding FrFT order: \\ $p = 0.7507$}} & {\makecell{Corresponding FrFT order: \\ $p = 0.8958$}} & {\makecell{Corresponding FrFT order: \\ $p \approx 1$}}\\
            ~ & {\makecell{Corresponding scale factor: \\ $s_2 = 2.6198$}} & {\makecell{Corresponding scale factor: \\ $s_2 = 6.1357$}} & {\makecell{Corresponding scale factor: \\ $s_2 = 242.1505$}}\\
            \cline{2-4}
            {\makecell{Field of view: \\ $51.2 \,um \times 51.2 \,um$}} & {\makecell{Field of view: \\ $134.13 \,um \times 134.13 \,um$}} & {\makecell{Field of view: \\ $314.88 \,um \times 314.88 \,um$}} & {\makecell{Field of view: \\ $12.40 \,mm \times 12.40 \,mm$}}\\
            {\makecell{Sampling number: \\ $256 \times 256$}} & {\makecell{Sampling number: \\ $256 \times 256$}} & {\makecell{Sampling number: \\ $256 \times 256$}} & {\makecell{Sampling number: \\ $256 \times 256$}}\\
            {\makecell{Pixel size: \\ $200 \,nm \times 200 \,nm$}} & {\makecell{Pixel size: \\ $0.52 \,um \times 0.52 \,um$}} & {\makecell{Pixel size: \\ $1.23 \,um \times 1.23 \,um$}} & {\makecell{Pixel size: \\ $48.43 \,um \times 48.43 \,um$}}\\
		\hline 
        \end{tabular}
	
    \end{table*}

    \subsection{Applications for coherent diffraction imaging}
    Coherent diffraction imaging (CDI) is a "lensless" technique for 2D or 3D reconstruction of the image of nanoscale structures such as nanocrystals \cite{pfeifer2006three,miao2015beyond}, potential proteins \cite{miao2011coherent,branden2019coherent}, and more \cite{marchesini2003coherent}.
    Due to the ill-posed nature of Fourier PR, existing CDI technologies mainly rely on the support constraint \cite{yang2021dynamic,kang2021single} or coded modulation conditions \cite{zhang2016phase} to achieve reconstruction. In this part, we unveil a novel imaging capability empowered by the proposed SFrFPR, \ie support-free coherent diffraction imaging. 
    To this end, we take practical experimental setups into account and perform numerical simulations to verify the proposed method. 

    The overall scheme of the proposed support-free CDI is illustrated in Fig. \ref{fig:overiew_cdi}. The incident X-ray energy is $5$ keV corresponding to the wavelength of $0.248$ nm. 
    The sample is a 3D particle of Bacterial RNA-free RNase P \cite{wilhelm2023bacterial}, whose volume data can be downloaded from the public protein data bank\footnote{https://www.rcsb.org/structure/8SSG.}.
    After the plane wave illuminates the particle, we use the projection approximation method \cite{morgan2010projection} to project the 3D volume data into the 2D exitwave represented as a phase object.
    Then through near-field diffraction, a single-shot intensity pattern of the FrFT measurement is collected by a detector. 
    For clarity, we follow the scalable sampling of the proposed FrFT measurement model and simulate the corresponding intensity pattern via the numerical integration of Fresnel Integral.
    More details of the optical setting can be found in Table. \ref{tab:exp_params}.
    Finally, the latent exitwave can be reconstructed from the single FrFT measurement using the proposed UNN-based method. It is worth emphasizing that the system does not require tight support of the object or additional physical equipment such as mask modulators.

\begin{figure}[t]
        \centering
        \small
        \setlength\tabcolsep{0.8pt}
        \begin{tabular}{cccc}
            {\makecell{FrFT Regime \\ ($d = 0.1 \,m$)}} & {\makecell{FrFT Regime \\ ($d = 0.25 \,m$)}} & {\makecell{Fourier Regime \\ ($d = 10 \,m$)}} & {Ground-Truth}\\
            \includegraphics[width=.24\linewidth,clip,keepaspectratio]{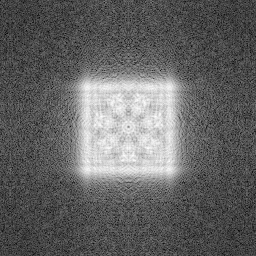} &
            \includegraphics[width=.24\linewidth,clip,keepaspectratio]{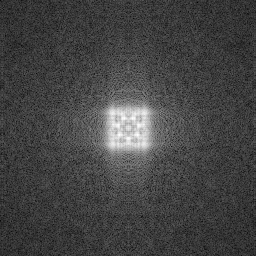} &
            \includegraphics[width=.24\linewidth,clip,keepaspectratio]{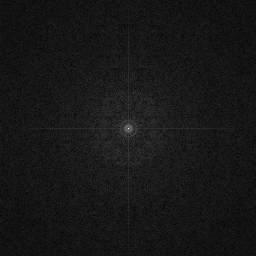} &
            \includegraphics[width=.24\linewidth,clip,keepaspectratio]{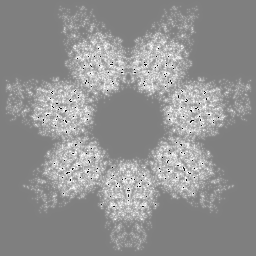} \\
            \includegraphics[width=.24\linewidth,clip,keepaspectratio]{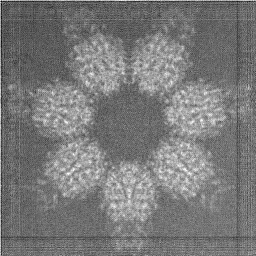} &
            \includegraphics[width=.24\linewidth,clip,keepaspectratio]{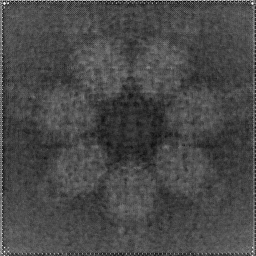} &
            \includegraphics[width=.24\linewidth,clip,keepaspectratio]{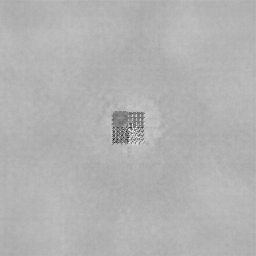} &
            \includegraphics[width=.24\linewidth,clip,keepaspectratio]{figures/SFCDI_200nm/2dexitwave.png} \\
        \end{tabular}
        \caption{The first three images in the top row show the diffraction patterns at $d$ values of 0.1 m, 0.25 m, and 10 m, respectively. Corresponding reconstructed objects from these patterns are shown below. The ground-truth object is listed in the last column.}
        \label{fig:scdi}
    \end{figure}

    We conducted three group experiments and collected intensity observations in different diffraction zones. The number of training epochs is 2000 for all reconstructions. Fig. \ref{fig:scdi} presents the reconstructed results when $d$ is 0.1 m, 0.25 m, and 10 m. It can be seen that the exitwave can be retrieved from the single intensity pattern in the FrFT regime while failing in the Fourier regime. Applying the intrinsic physical constraints of SFrFPR, we mitigate the inherent ambiguities of the reconstruction and achieve the support-free CDI technique. In particular, the proposed method is demonstrated to be robust to the inconsistency between the FrFT-based measurement model and the numerical integration model. All results support the potential of the proposed method in real experiments.

    \section{Conclusion}
    \label{sec:conclusion}    
    In this work, we have tackled the problem of single-shot phase retrieval from a fractional Fourier transform perspective. Specifically, we introduced the FrFT to resolve the perennial issue of numerical inaccuracies arising from the sampling constraints associated with the discretized transfer function involved in the Fresnel diffraction integral.
    Consequently, the FrFT-based measurement model, presented herein, emerges as a versatile solution that aptly addresses wave propagation scenarios spanning both short and long distances.
    In addition, we have embraced a self-supervised reconstruction framework that combines the inherent constraints of the FrFT measurement with untrained neural network priors, relaxing the previous conditions of oversampled or multiple measurements in the Fourier domain. Correspondingly, we demonstrate the rationale behind the proposed SFrFPR paradigm from the perspective of fractional Fourier time-frequency representation. Through numerical simulations, the results manifest a profound superiority of the single FrFT measurement in comparison to its Fourier transform counterparts. Moreover, the SFrFPR paradigm, as proposed, unveils the potential to revolutionize imaging paradigms, particularly in support-free coherent diffraction imaging. In the future, we believe the single-shot imaging capability of the proposed SFrFPR will have the potential to study dynamic processes in materials and biological science utilizing pulsed sources, such as X-ray free-electron lasers.
	
    
    %
    
    
    



    \ifCLASSOPTIONcaptionsoff
    \newpage
    \fi

    
    
    \bibliographystyle{IEEEtran}
    \bibliography{IEEEabrv}
\end{document}